\documentclass{article}

\usepackage[dblblindworkshop, final]{neurips_2026}

\workshoptitle{AI for Science}

\usepackage[utf8]{inputenc} %
\usepackage[T1]{fontenc}    %
\usepackage{hyperref}       %
\usepackage{url}            %
\usepackage{booktabs}       %
\usepackage{amsfonts}       %
\usepackage{nicefrac}       %
\usepackage{microtype}      %
\usepackage{xcolor}         %

\usepackage{amsmath}        %
\usepackage{graphicx}       %
\usepackage{placeins}       %
\usepackage{amssymb}
\usepackage{array}

\usepackage{cleveref}

\title{Beyond PINNs: Boundary-Enriched Holomorphic Networks for Laplace Problems}

\author{%
    Andrei Vlad Nicula, \quad Kalp Shah, \quad
    Robert-\c{S}tefan Sofroni, \quad Mihnea Andrei Bârsan,\\
    \textbf{Agneya Bhardwaj}, \quad \textbf{Jan Jelínek}, \quad
    \textbf{Chris Hoes}, \quad \textbf{Remco Hogerwerf}, \quad \textbf{Finn Govers},\\
    \textbf{Storm Koper}, \quad
    \textbf{Gerrit Janssen}, \quad \textbf{Stef de Wildt}\\
    University of Amsterdam\\
    \texttt{\{andrei.nicula, kalp.shah, robert.sofroni, mihnea.barsan,}\\
    \texttt{agneya.bhardwaj, jan.jelinek, chris.hoes, remco.hogerwerf, finn.govers, }\\
    \texttt{storm.koper, gerrit.janssen.janssen,stef.de.wildt\}@student.uva.nl}
    \\
    \textbf{Alejandro Valverde Mahou}\\
    Technical University of Denmark\\
    \texttt{avama@dtu.dk}
}

\begin{document}

\maketitle

\begin{abstract}
Physics-informed neural networks (PINNs) solve partial differential equations by penalizing their residual, but this formulation can be difficult to optimize for solutions with singular behaviour. We focus specifically on the two-dimensional Laplace equation, whose strong analytical structure allows the governing equation to be enforced directly in the neural representation. We compare a hybrid harmonic--neural model with a holomorphic neural network whose output is harmonic by construction and therefore requires only boundary training. We further enrich the holomorphic model with singular functions chosen according to the interior angle and boundary conditions at each junction of boundary segments. On six singular benchmarks, the enriched model is the most accurate of the neural models we compare. On the four with a closed-form solution, however, that solution lies in the span of the enrichment terms, and a least-squares fit of these terms alone is exact to rounding error. On Motz's problem and the cracked beam, the neural core reduces the relative error of the recovered leading singular coefficient from about $10^{-2}$, for a least-squares fit with the same four terms, to about $10^{-6}$, and on four variants whose solution has a smooth part outside the span, the network is $4\times10^{4}$ to $2\times10^{7}$ times more accurate than a fit of its singular terms alone; classical least-squares fits with more singular terms, or with harmonic polynomials and rational poles, are more accurate still. Our results demonstrate how exploiting problem-specific analytical structure can provide an alternative to base and residual PINNs for challenging scientific computing problems.

\end{abstract}

\section{Introduction}
\label{sec:introduction}

A common way to incorporate physical laws into neural networks is through physics-informed neural networks (PINNs), which enforce the governing equations through the training objective rather than through an explicit discretization \citep{raissi2019pinns}. The solution is represented as a continuous network whose parameters are optimized so that the PDE residual at interior collocation points and the mismatch against the prescribed boundary data are simultaneously small, with derivatives supplied by automatic differentiation. This makes PINNs mesh-free, differentiable end-to-end, and straightforward to integrate into deep-learning pipelines, motivating their use for irregular geometries and inverse problems where data and physics must be combined \citep{luo2025review}.

The difficulty is that this objective is hard to optimize. Its terms compete:
\citet{wang2021understanding} identify a gradient pathology in which the
residual and boundary contributions differ by orders of magnitude, and \citet{wang2022ntk} recover the same effect through the neural tangent kernel. Networks also carry a spectral bias towards smooth, low-frequency functions \citep{rahaman2019spectralbias}, which is the wrong bias for solutions with sharp gradients. Both are at their worst on elliptic problems with mixed boundary conditions, where the solution loses regularity at corners and wherever the boundary condition changes type or value \citep{grisvard_elliptic_1985}.

Rather than modifying the optimization, we exploit the analytical structure of the PDE itself. Laplace's equation, $\Delta u=0$, is particularly well suited to this approach: its solutions form a linear space, are controlled by their boundary data, and satisfy the maximum principle.

We therefore explore two increasingly structured ways of incorporating the
properties of Laplace's equation into the neural representation. The first retains the PINN framework but writes the solution as a finite harmonic expansion plus a learned residual, $u=u_h+r_\theta$, so that the network only has to learn the correction that the exact harmonic expansion cannot represent.

The second approach goes further and removes the interior residual entirely.
A network built from complex affine layers and holomorphic activations is
holomorphic for every value of its weights, and its real part is therefore
harmonic. The governing equation is thus satisfied by construction, leaving
only the boundary conditions to be fitted. We refer to this as the
\emph{plain holomorphic network}, following related harmonic and holomorphic
neural-network constructions \citep{ghosh_harmonic_2023, calafa_physics-informed_2024}.

However, exact harmonicity alone does not resolve the difficulty of singular
solutions. We therefore enrich the holomorphic representation with singular
terms derived from the boundary specification, while preserving exact
harmonicity. The corresponding singular exponents can be learned rather than
fixed in advance; we refer to this as the \emph{enriched holomorphic network}.

Together, these models let us study the effect of progressively incorporating analytical structure into the representation. We compare them with standard neural and harmonic baselines on Laplace problems with singularities induced by re-entrant corners, jumps in the data, and Dirichlet--Neumann junctions, and we compare the enriched network with a least-squares fit of its singular terms alone, which separates the contributions of the neural core and the enrichment terms. On the classical closed-form benchmarks, the exact solution is itself a finite combination of these terms, so the fit is as accurate as the network, making these benchmarks serve as consistency checks. On Motz's problem, the cracked beam, and four variants with a smooth part outside the span of the terms, the core has to supply the rest, and the enriched network is four to seven orders of magnitude more accurate than the fit of its terms alone, although classical fits with more terms are more accurate still.

\section{Previous work}
\subsection{Physics-informed neural networks}
Physics-informed neural networks (PINNs) \citep{raissi2019pinns} incorporate
the governing equations into the training objective through PDE residuals.
Much of the subsequent work has focused on improving the optimization of this
residual-based formulation. Adaptive loss weighting has been used to balance
the different components of the objective
\citep{kendall2018uncertainty, heydari2019softadapt, wang2022ntk}, while
adaptive collocation strategies modify where the PDE residual is evaluated
\citep{wu2023sampling, mcclenny2023self}. Other approaches modify the network
representation, for example through periodic activations or Fourier features
\citep{sitzmann2020siren, liu2025fourier}, or replace the standard MLP
backbone with architectures such as Kolmogorov--Arnold networks
\citep{liu2024kan,dzimah2026unif}.

These approaches retain the residual-based formulation of PINNs; we instead build harmonicity into the representation, so that no interior penalty is needed.

\subsection{Structure-preserving representations for Laplace problems}

A direct way to exploit the structure of Laplace's equation is to represent the solution as a finite expansion of harmonic functions,
\begin{equation}
u_h(x,y)=\sum_k c_k\phi_k(x,y),
\qquad
\Delta\phi_k=0,
\label{eq:harmonic}
\end{equation}
for which the PDE is satisfied exactly by linearity. Such representations arise classically from separation of variables on simple geometries or from harmonic kernels and fundamental solutions centred outside the domain \citep{axler2001harmonic, Schaback2008kernels}. The resulting coefficients can be fitted using boundary data alone. Their main limitation is approximation capacity: the harmonic function space is infinite-dimensional, so a finite dictionary must be chosen in advance and may require many terms to represent irregular geometries or solutions with reduced regularity.

The same principle underlies classical boundary-based numerical methods such as boundary element methods, the method of fundamental solutions, and the lightning Laplace solver, which build approximations from functions satisfying the governing equation in the interior and thereby reduce the problem to determining parameters from boundary information \citep{fairweather_method_1998, gopal_solving_2019}. Neural approaches extend the same principle, parameterizing the admissible family with a network rather than with a fixed analytical basis.

\subsection{Holomorphic neural networks}

In two dimensions, complex analysis provides a stronger structure-preserving representation: the real part of a holomorphic function is harmonic by the Cauchy--Riemann equations, so a network assembled from complex affine maps and holomorphic activations parameterizes harmonic functions without the Laplace residual in its objective.

This construction has recently been explored in neural PDE solvers. \citet{ghosh_harmonic_2023} use complex-valued networks to represent harmonic functions through their real part, while the physics-informed holomorphic neural networks of \citet{calafa_physics-informed_2024} extend the idea to plane linear elasticity using holomorphic potentials, and study approximation properties and initialization strategies for such networks.

\subsection{Singularity-aware representations}
\label{sec:singular_rep}

Exact satisfaction of the governing equation does not, however, guarantee that the chosen representation can efficiently approximate the solution. Elliptic problems with corners or mixed boundary conditions can exhibit reduced regularity at points where boundary segments meet or where the type of boundary condition changes. Locally, such solutions can contain fractional-power terms whose exponents depend on the local geometry and the boundary conditions, a term proportional to the polar angle where Dirichlet data jump, and $r\log r$ terms where an exponent resonates with the data, as at a flux jump on a straight edge or at a right-angle Dirichlet--Neumann corner with nonzero flux~\citep{grisvard_elliptic_1985}.

Singular-function enrichment is a classical response to this problem. Finite- and boundary-element methods augment their approximation spaces with analytically known singular functions to capture local behaviour that would otherwise require a large number of regular basis functions \citep{fix_use_1973, li_singularities_2000}. The same principle has recently been applied to neural PDE solvers. SEPINN \citep{hu_solving_2024}, for example, augments a real-valued PINN with singular functions associated with polygonal corners, with their exponents determined analytically from the geometry, while retaining the conventional interior PDE residual. \citet{calafa_solving_2025} similarly enrich holomorphic neural networks with the square-root term from Williams' expansion to represent the stress singularity at a crack tip, with the exponent fixed by the crack geometry. Our exponents are likewise set analytically from the angle and boundary conditions at each junction, and in every run we report they are pinned there or start there, apart from the recovery runs of \cref{app:lambda}; we differ from SEPINN in removing the interior residual altogether, and from \citet{calafa_solving_2025} in treating scalar Laplace problems with mixed Dirichlet--Neumann data, with terms selected per junction from \cref{tab:dictionary} rather than from Williams' crack-tip expansion.

\subsection{Relation to classical singular solvers}

The enriched network is a neural counterpart of singular-function (Trefftz) least-squares methods, which expand the solution in the same local series and produced the reference values of $D_0$ we use~\citep{li_singularities_2000,lu_highly_2004}, and of lightning and AAA-least-squares solvers, which place poles of a rational function near the singular points and converge root-exponentially without knowing the exponents~\citep{gopal_solving_2019, costa_aaa-least_2023}. A linear least-squares fit of our own singular terms, with more terms and, where needed, harmonic polynomials and AAA-located poles, matches or exceeds the enriched network in seconds rather than tens of minutes to hours (\cref{sec:results}), so we claim no accuracy or speed advantage over these methods. What the neural formulation offers is a nonlinear, exactly harmonic model trained with the same gradient-based tools as other networks, into which classical singular terms can be inserted directly. Its use inside larger learning pipelines, such as inverse or parametric problems, is not demonstrated here.
\section{Method}
\label{sec:method}

We consider three representations: (i) a harmonic expansion plus a PINN residual, which encodes harmonicity only in a fixed component, (ii) a holomorphic neural network, harmonic for every set of parameters, and (iii) an enriched holomorphic network that additionally represents the local singular structure at boundary junctions.

\subsection{Problem formulation}

We study the two-dimensional Laplace equation on a bounded domain
$\Omega \subset \mathbb{R}^2$ with mixed \emph{Dirichlet--Neumann} boundary conditions~\citep{evans2010pde, gilbarg2001elliptic}.
The boundary is partitioned as $\partial\Omega=\Gamma_D\cup\Gamma_N$, where
$\Gamma_D\cap\Gamma_N=\emptyset$. The solution $u:\Omega\rightarrow\mathbb{R}$ satisfies
\begin{equation}
\label{eq:laplace_bvp}
\Delta u=0 \ \text{in } \Omega,
\qquad
u=g_D \ \text{on } \Gamma_D,
\qquad
\nabla u\cdot n=g_N \ \text{on } \Gamma_N ,
\end{equation}
where $g_D$ and $g_N$ are the prescribed \emph{Dirichlet} and \emph{Neumann} data and $n$ the outward unit normal.

\subsection{Hybrid harmonic-neural model}

Our model decomposes the solution into a harmonic component and a learned residual:
\begin{equation}
\label{eq:hybrid}
u_{\theta}(x,y)=u_h(x,y)+r_{\theta}(x,y),
\qquad
\Delta u_\theta=\Delta r_\theta ,
\end{equation}
where $u_h$ is the finite harmonic expansion in \cref{eq:harmonic}, and $r_\theta$ is parameterized by a PINN. The harmonic component therefore satisfies the Laplace equation by construction, and the network is responsible only for the correction that a finite expansion cannot represent.
The hybrid loss is:
\begin{equation}
\label{eq:hybrid_loss}
\mathcal{L}_{\mathrm{hybrid}}
=
\frac{1}{|\mathcal{X}_{\Omega}|}
\sum_{x_i\in\mathcal{X}_{\Omega}}
|\Delta r_\theta(x_i)|^2
+
w_{\mathrm{bc}}
\left(\mathcal{L}_{D}^{h+r}+\mathcal{L}_{N}^{h+r}\right),
\end{equation}
where the two boundary terms are those of the PINN objective, evaluated on the
full prediction $u_h + r_\theta$ rather than on the network alone.

The hybrid model still trains an interior residual and still needs a dictionary of harmonic functions fixed in advance, which limits its performance and usability.

\subsection{Holomorphic functions}
\label{sec:holomorphic}

We identify the two-dimensional domain $\Omega\subset\mathbb{R}^2$ with a subset of the complex plane through $z=x+\mathrm{i}y$, and represent the real-valued solution as the real part of a complex-valued function,
\begin{equation}
u(x,y)=\operatorname{Re} f(z).
\label{eq:holo-ansatz}
\end{equation}
If $f$ is holomorphic in $\Omega$, its real part is harmonic by the Cauchy--Riemann equations. We therefore construct $f$ as a neural network that is holomorphic by construction. Each layer consists of a complex affine map followed by a componentwise exponential activation, $z \mapsto \exp(Wz+b)$.

Both the affine map and the exponential are entire functions, and their composition is therefore holomorphic for every value of the network parameters. The choice of exponential provides a non-polynomial entire activation, as by Liouville's theorem, any bounded entire activation would have to be constant. Consequently, the predicted solution in
\cref{eq:holo-ansatz} satisfies $\Delta u_\theta = 0$ throughout the domain, independently of the network parameters and of the state of training. The Laplace residual therefore does not need to be evaluated or included in the training objective. The complex input is affinely normalized as $\hat z=(z-z_0)/s$, where $z_0$ is the centre of the domain and $s$ its characteristic radius. This transformation preserves holomorphy.

The boundary conditions are then imposed directly on the harmonic representation. \emph{Dirichlet} data are evaluated from \cref{eq:holo-ansatz}.

For \emph{Neumann} data, the Cauchy--Riemann equations give
$f'(z)=u_x-\mathrm{i}u_y$. Hence, at a boundary point with outward unit normal $n=(n_x,n_y)$,
\begin{equation}
\frac{\partial u}{\partial n}
=
\operatorname{Re}\left[(n_x+\mathrm{i}n_y)f'(z)\right].
\label{eq:neumann-head}
\end{equation}
Thus the same holomorphic representation provides both the function value and its normal derivative, allowing mixed \emph{Dirichlet--Neumann} problems to be trained using boundary data alone.

The resulting objective contains only the two boundary misfits,
\begin{equation}
    \mathcal{L} = \frac{w_D}{|X_D|} \sum_{x_i \in X_D} \left(u_\theta(x_i) - g_D(x_i)\right)^2
    + \frac{w_N}{|X_N|} \sum_{x_i \in X_N} \left(\partial_n u_\theta(x_i) - g_N(x_i)\right)^2 ,
    \label{eq:boundary-loss}
\end{equation}
where $X_D$ and $X_N$ denote sampled points on the \emph{Dirichlet} and \emph{Neumann} parts of the boundary. There is no interior residual and hence no weighting between a PDE residual and the boundary data. Throughout the experiments, we use $w_D=w_N=1$.

Exact harmonicity also relates the interior error to the boundary misfit. Writing $u_\theta$ for the trained model, $u^\star$ for the exact solution, and $g$ for the prescribed data, the error $e = u_\theta - u^\star$ of a problem with Dirichlet data on every segment is itself harmonic, since both terms are harmonic and the Laplacian is linear. If $e$ is bounded in $\Omega$, the maximum principle, applied to $e$ and $-e$, gives
\begin{equation}
  \sup_{\Omega}\,|u_\theta-u^\star| \;\le\; M:=\sup_{\partial\Omega\setminus E}\,|u_\theta-g| ,
  \label{eq:estimate}
\end{equation}
where $E$ is the finite set of junctions, at which $g$ may jump. Jumps in the data are therefore allowed, provided the model stays bounded near them, and every enrichment term of \cref{sec:enrichment} does, which is why the coefficient of the value-jump term is restricted there to be imaginary. The right-hand side needs no reference solution, whereas a soft-constraint PINN admits no such relation, since its error is not harmonic. We approximate $M$ by the largest misfit over a finite sample of boundary points, excluding small discs around the singular sites (\cref{app:holomorphic_evaluation}), and call the result $\hat M$ the \emph{sampled maximum-principle estimate}. Being a maximum over part of the boundary, $\hat M\le M$ in exact arithmetic, and it omits the misfit between samples and near the sites: it estimates the interior error but does not bound it. \Cref{sec:results,app:holomorphic_evaluation} report how closely it matches.

\subsection{Enrichment}
\label{sec:enrichment}

We augment the holomorphic network with singular terms associated with boundary junctions. The boundary is provided as a sequence of straight or circular segments, each with a boundary condition and its prescribed data, and our compiler inspects every junction $z_c$ between adjacent segments, whose local behaviour is determined by the interior angle $\omega$ and the conditions on the two sides. In the local polar coordinate $\zeta=z-z_c=re^{\mathrm{i}\theta}$, the Laplace equation on the corresponding wedge admits separated solutions of the form $
r^\lambda \left(A\cos(\lambda\theta)+B\sin(\lambda\theta)\right)$.

The boundary conditions on the two sides fix the admissible exponents $\lambda$; for non-integer $\lambda$ the gradient behaves as $r^{\lambda-1}$ and is singular when $\lambda<1$. At a \emph{Dirichlet--Neumann} junction, for example, $\lambda_k=(2k-1)\pi/(2\omega)$, $k=1,2,\ldots$, and the mode $r^{\lambda_k}\sin(\lambda_k\theta)$ is the real part of $-\mathrm{i}\,\zeta^{\lambda_k}$. The data can also add two logarithmic terms: a Dirichlet value jump adds the angle $\theta=\operatorname{Im}\operatorname{Log}\zeta$, and a nonzero flux where an exponent is an integer, as at a right-angle \emph{Dirichlet--Neumann} corner, adds a term $\propto r\ln r$, represented by $\zeta\operatorname{Log}\zeta$. \Cref{tab:dictionary} lists the term added in each configuration.

\begin{table}[t]
  \caption{The enrichment dictionary for a junction $z_c$ of two straight segments with piecewise-constant data, by the conditions on its two sides (D \emph{Dirichlet}, N \emph{Neumann}). Here $\zeta=z-z_c$, $r=|\zeta|$, $\theta$ is the angle from the first segment (the Dirichlet one at a D--N junction), $\omega$ the interior angle, $h,h_a,h_b$ outward fluxes and $k=1,2,\dots$ the ladder index; only non-integer exponents are added. A junction gains the terms of every row it matches, each with a trainable complex coefficient (real for $\mathrm{i}\operatorname{Log}\zeta$). Every term is holomorphic in the domain, so the model stays exactly harmonic.}
  \label{tab:dictionary}
  \centering
  \footnotesize
  \begin{tabular}{@{}lllll@{}}
    \toprule
    Sides & Condition & Solution near $z_c$ & $|\nabla u|$ & Term added \\
    \midrule
    D--N & any $\omega$ & $r^{\lambda}\sin(\lambda\theta)$, $\lambda=\tfrac{(2k-1)\pi}{2\omega}$ & $r^{\lambda-1}$ & $\zeta^{\lambda}$ \\
    D--N & $\cos\omega=0$, $h\neq0$ & $\propto h\,r\ln r$ & $|\ln r|$ & $\zeta\operatorname{Log}\zeta$ \\
    \addlinespace
    D--D & value jump $\Delta g$ & $(\Delta g/\omega)\,\theta$ & $r^{-1}$ & $\mathrm{i}\operatorname{Log}\zeta$ \\
    D--D & $\pi/\omega\notin\mathbb{Z}$ & $r^{\lambda}\sin(\lambda\theta)$, $\lambda=k\pi/\omega$ & $r^{\lambda-1}$ & $\zeta^{\lambda}$ \\
    \addlinespace
    N--N & $\pi/\omega\notin\mathbb{Z}$ & $r^{\lambda}\cos(\lambda\theta)$, $\lambda=k\pi/\omega$ & $r^{\lambda-1}$ & $\zeta^{\lambda}$ \\
    N--N & $\sin\omega=0$, $h_b+h_a\cos\omega\neq0$ & $\propto r\ln r$ & $|\ln r|$ & $\zeta\operatorname{Log}\zeta$ \\
    \addlinespace
    \multicolumn{2}{@{}l}{Otherwise} & analytic & bounded & none \\
    \bottomrule
  \end{tabular}
\end{table}
The enriched model adds the compiled terms, which supply the part of the solution that an entire function approximates poorly, to a holomorphic core $N_0$,
\begin{equation}
    f(z) = N_0(\hat z) + \sum_{c \in \mathcal{S}} \sum_{k=1}^{m_c} \beta_{c,k} E_{c,k}(\zeta_c),
    \label{eq:enriched}
\end{equation}
where $\mathcal{S}$ is the compiled set of sites, $E_{c,k}$ is the term that \cref{tab:dictionary} assigns to site $c$ at ladder index $k$, and
$\beta_{c,k} \in \mathbb{C}$ is trainable. We use $m_c=1$ except on Motz's problem, its variant, and the cracked beam, where $m_c=4$. The mode of a value jump has a single real amplitude, that of the angle $\theta=\operatorname{Im}\operatorname{Log}\zeta$, so its term is $\mathrm{i}\,b\operatorname{Log}\zeta$ with $b\in\mathbb{R}$: a real part would add $\operatorname{Re}\beta\,\ln r$, which is unbounded at the site, and \cref{eq:estimate} would no longer apply. Every enrichment term therefore has a bounded real part. The coefficients start at zero, so training begins from the bare core, and site positions, branch-cut directions, chosen so that every cut stays outside the domain (\cref{app:holomorphic_implementation}), and interior angles are fixed by the specification and never trained.

Because $\zeta^{\lambda}$ is holomorphic for every $\lambda > 0$, the exponents can also be trained without any effect on harmonicity (\cref{app:lambda}). We pin them where a published coefficient is recovered; \cref{app:holomorphic_training} lists the setting of every run.

\section{Experiments}

\subsection{Problem definitions}
\label{sec:problems}
Two families of problems are used for the enriched holomorphic model. The first is a set of six singular benchmarks, each with a single singular site. Four have a closed-form solution and are scored by relative $L^2$ error against it: two circular sectors, an L-shaped domain, and a slit disk, which are standard test domains whose solution is the leading wedge mode $r^{\lambda}\sin(\lambda\theta)$ of the corresponding opening angle \citep{grisvard_elliptic_1985}. This mode is exactly the leading term that \cref{tab:dictionary} assigns to the site, and the only one the model carries there, so these four problems are consistency checks: they test whether training switches the core off and recovers one coefficient, with the exponent starting at its exact value, and \cref{tab:all_models} adds a least-squares fit of that term alone. The other two have no closed form, and the literature reports instead the leading coefficient $D_0$ of the expansion around the singularity. These are Motz's problem, introduced by \citet{motz_treatment_1947} and used as a prototype for singularity treatments \citep{li_singularities_2000}, and the cracked beam, for which we take the domain and boundary conditions of \citet{lu_highly_2004} so that their published coefficient applies. On these two problems, the model carries four ladder terms at the site, with pinned exponents (\cref{tab:benchmarks}), as does the variant of Motz's problem in \cref{tab:all_models}. Every other problem has one term per site.

\Cref{tab:all_models} also includes four variants that keep the geometry, boundary types, and singular site of a parent problem but add a smooth harmonic function, so that the solution has a part outside the span of the singular terms (\cref{app:holomorphic_benchmarks}).

The second is a suite of 24 randomly sampled problems, twelve on the square and twelve on the disk, four in each of three families: \emph{Dirichlet} data with value jumps, mixed \emph{Dirichlet} and \emph{Neumann} data, and configurations with many segments. These have no closed form in general, so they are scored by the boundary loss, and the disk instances with Dirichlet data on every segment, whose solutions are sums of harmonic measures of arcs, also by relative $L^2$ error. Where every segment carries Dirichlet data, we also report the sampled maximum-principle estimate $\hat M$ of \cref{sec:holomorphic}, which \cref{sec:results} compares with the interior error.
\Cref{app:holomorphic_suites} shows the suite and the singular sites the compiler emits on each instance.

\subsection{Results}
\label{sec:results}

\Cref{fig:pairs} in \cref{app:holomorphic_suites} compares the plain and the enriched holomorphic network over the sampled suite. Both are trained with the same budget and restarts, and differ only in the compiled singular terms, whose exponents are fixed at their values in \cref{tab:dictionary}. The enriched model lowers the boundary loss in every pair by a factor of at least $13$. On the Dirichlet-jump families, all 24 enriched runs reach a boundary loss below $10^{-6}$, with median improvements of $3.0\times10^{14}$ (disk) and $1.1\times10^{8}$ (square). On the families with Neumann data, the median improvement is $5.0\times10^{2}$ to $3.6\times10^{3}$, and 17 of 48 runs fall below $10^{-6}$. \Cref{app:holomorphic_suites} gives the results by family.

\begin{figure}[ht]
    \centering
    \includegraphics[width=.75\linewidth]{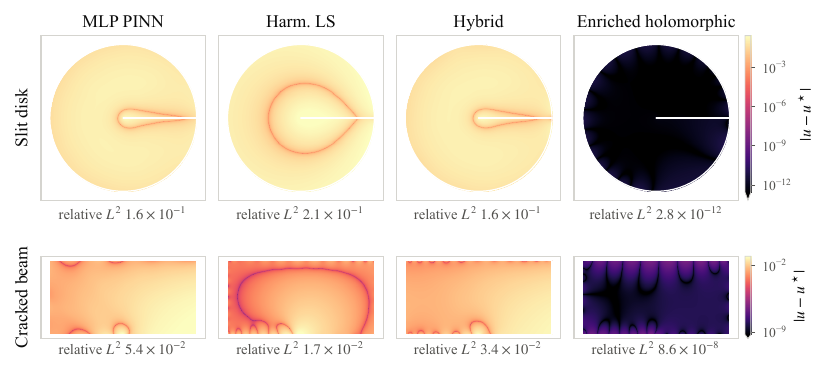}
    \caption{Error $|u-u^\star|$ of the MLP PINN, the harmonic least-squares fit, the hybrid harmonic--neural model and the enriched holomorphic network on the \emph{Slit disk} (top) and the \emph{Cracked beam} (bottom, against the series reference), on one logarithmic scale per problem.}
    \label{fig:results}
\end{figure}

On the 24 Dirichlet-jump pairs of \cref{fig:pairs}, enrichment lowers $\hat M$ by a median factor of $2.8\times10^{4}$ (square) and $1.1\times10^{8}$ (disk), and $\hat M$ tracks the interior error of the enriched model: at distances of at least $10^{-6}$ from the jumps, the interior maximum error lies within $1.8\%$ (disk) and $11\%$ (square) of $\hat M$. The same holds where the data are continuous: on the two sectors, the L-shape and the slit disk, the interior sup error and $\hat M$ agree to within $2\times10^{-4}$ in 41 of the 48 runs behind \cref{tab:all_models}, and in the other seven both are at rounding level. \Cref{fig:results} shows the error of the four models on the \emph{Slit disk} and the \emph{Cracked beam}. Only the harmonic fit concentrates its error at the singular site. The MLP PINN and the hybrid model spread theirs over the whole domain, so the singularity is not their only source of error. The enriched model's error is of order $10^{-12}$ on the slit disk and $10^{-7}$ on the cracked beam, the latter against the series reference. \Cref{app:holomorphic_benchmarks} details the benchmarks and their references.

\begin{table}[ht]
  \caption{Errors on the named singular benchmarks and four out-of-span variants (\cref{sec:problems}): median over three seeds at the best of the four training budgets, chosen for each method and metric. The best entry in each row is bold; entries below $10^{-15}$ are at rounding level and are bold alike. Harmonic fits a smooth harmonic basis and Enr.\ LS the enriched model's singular terms, both by least squares without a network; with forty ladder terms instead of four, Enr.\ LS recovers $D_0$ to below $10^{-13}$. The series reference reproduces the published $D_0$ of \citet{lu_highly_2004} to $10^{-15}$; only the $D_0$ rows are independent of the enriched model's basis.}
  \label{tab:all_models}
  \centering
  \footnotesize
  \setlength{\tabcolsep}{4pt}
  \begin{tabular}{lrrrrrr}
    \toprule
    & & & & \multicolumn{2}{c}{Holomorphic} & \\
    \cmidrule(lr){5-6}
    Problem & \multicolumn{1}{c}{MLP} & \multicolumn{1}{c}{Harmonic} & \multicolumn{1}{c}{Hybrid} & \multicolumn{1}{c}{Plain} & \multicolumn{1}{c}{Enriched} & \multicolumn{1}{c}{Enr.\ LS} \\
    \midrule
    \multicolumn{7}{l}{\emph{Closed-form solution, relative $L^2$ error}} \\
    Sector $3\pi/2$ & $2.5\times 10^{-2}$ & $3.2\times 10^{-2}$ & $3.0\times 10^{-2}$ & $7.7\times 10^{-4}$ & $3.3\times 10^{-12}$ & $\mathbf{5.5\times 10^{-16}}$ \\
    Sector $7\pi/4$ & $5.0\times 10^{-2}$ & $1.1\times 10^{-1}$ & $5.2\times 10^{-2}$ & $1.9\times 10^{-2}$ & $\mathbf{1.4\times 10^{-16}}$ & $\mathbf{9.9\times 10^{-16}}$ \\
    L-shape & $1.5\times 10^{-2}$ & $1.8\times 10^{-2}$ & $1.6\times 10^{-2}$ & $7.5\times 10^{-4}$ & $\mathbf{2.9\times 10^{-16}}$ & $\mathbf{5.3\times 10^{-16}}$ \\
    Slit disk & $1.6\times 10^{-1}$ & $2.1\times 10^{-1}$ & $1.6\times 10^{-1}$ & $3.2\times 10^{-1}$ & $2.8\times 10^{-12}$ & $\mathbf{2.0\times 10^{-16}}$ \\
    \midrule
    \multicolumn{7}{l}{\emph{Series reference, relative $L^2$ error}} \\
    Motz & $1.1\times 10^{-1}$ & $1.7\times 10^{-2}$ & $5.0\times 10^{-2}$ & $5.4\times 10^{-4}$ & $\mathbf{2.5\times 10^{-8}}$ & $2.7\times 10^{-3}$ \\
    Cracked beam & $5.4\times 10^{-2}$ & $1.7\times 10^{-2}$ & $3.4\times 10^{-2}$ & $6.4\times 10^{-4}$ & $\mathbf{8.6\times 10^{-8}}$ & $1.1\times 10^{-2}$ \\
    \midrule
    \multicolumn{7}{l}{\emph{Leading coefficient, relative error of $D_0$}} \\
    Motz & $1.6\times 10^{-1}$ & $5.0\times 10^{-2}$ & $6.2\times 10^{-2}$ & $4.2\times 10^{-4}$ & $\mathbf{8.2\times 10^{-7}}$ & $1.1\times 10^{-2}$ \\
    Cracked beam & $6.0\times 10^{-2}$ & $5.1\times 10^{-2}$ & $3.0\times 10^{-2}$ & $5.9\times 10^{-4}$ & $\mathbf{2.0\times 10^{-6}}$ & $2.8\times 10^{-2}$ \\
    \midrule
    \multicolumn{7}{l}{\emph{Out-of-span variants, relative $L^2$ error}} \\
    L-shape $+\operatorname{Re}e^{z}$ & $7.5\times 10^{-3}$ & $7.6\times 10^{-3}$ & $6.6\times 10^{-3}$ & $2.8\times 10^{-4}$ & $\mathbf{1.5\times 10^{-8}}$ & $1.4\times 10^{-1}$ \\
    Pure mode $+\operatorname{Re}e^{z}$ & $5.3\times 10^{-3}$ & $4.9\times 10^{-3}$ & $3.0\times 10^{-3}$ & $4.4\times 10^{-4}$ & $\mathbf{1.4\times 10^{-8}}$ & $2.3\times 10^{-1}$ \\
    Sector $3\pi/2$ + pole & $1.3\times 10^{-1}$ & $5.1\times 10^{-2}$ & $5.0\times 10^{-2}$ & $6.3\times 10^{-4}$ & $\mathbf{4.6\times 10^{-6}}$ & $9.3\times 10^{-1}$ \\
    Motz + pole & $4.4\times 10^{-2}$ & $1.1\times 10^{-2}$ & $3.4\times 10^{-2}$ & $8.3\times 10^{-4}$ & $\mathbf{5.4\times 10^{-6}}$ & $2.0\times 10^{-1}$ \\
    \bottomrule
  \end{tabular}
\end{table}

Among the networks and baselines of \cref{tab:all_models}, the enriched holomorphic network is the most accurate on all six named problems: five to fifteen orders of magnitude below the MLP PINN, the harmonic fit, and the hybrid model, which all lie between $1.5\times10^{-2}$ and $2.1\times10^{-1}$, and eight to fourteen orders below the plain network on the closed-form problems and about four on Motz and the cracked beam. It recovers $D_0$ to $8.2\times10^{-7}$ and $2.0\times10^{-6}$. On the closed-form problems, the solution is the single enrichment term, and fitting that term alone by least squares is exact to rounding in about a second, against $8\times10^{2}$ to $9\times10^{3}$~s per network run. On Motz and the cracked beam the four terms alone recover $D_0$ only to $1.1\times10^{-2}$ and $2.8\times10^{-2}$, so at a matched number of terms the core gains about four orders. With forty terms, however, the same fit, a least-squares form of the collocation Trefftz method behind the published values~\citep{li_singularities_2000,lu_highly_2004}, reaches $1.6\times10^{-14}$ and $5.3\times10^{-14}$ with a linear solve of about 15~ms.

On the four variants, the solution has a smooth part outside the span of the singular terms, and a least-squares fit of those terms alone leaves a relative $L^2$ error of $1.4\times10^{-1}$ to $9.3\times10^{-1}$. With the same terms, the enriched network reaches $1.4\times10^{-8}$ to $5.4\times10^{-6}$, $3.7\times10^{4}$ to $1.6\times10^{7}$ times below that fit and $1.4\times10^{2}$ to $3.1\times10^{4}$ times below the plain network, so here the core supplies what the terms cannot. It is still not competitive with classical solvers: adding harmonic polynomials of degree~16 to the singular terms, and AAA-located poles~\citep{costa_aaa-least_2023} on the two pole variants, gives a least-squares fit of $6\times10^{-15}$ to $3\times10^{-14}$ in about 7~s or less.

\section{Conclusion}

We compared two ways of moving known structure of the Laplace problem from the
training objective into the model: a hybrid model that fixes part of the
hypothesis space and learns the rest, and holomorphic networks that fix all of
it and fit boundary data alone. The enriched holomorphic network is the most accurate of the neural models we tested. On the closed-form benchmarks, the singular terms alone already give the solution. On Motz's problem, the cracked beam and the smooth variants they do not, and the neural core supplies the rest. Classical least-squares fits with more singular terms, harmonic polynomials or poles nevertheless match or exceed the network on every named benchmark and variant, at a fraction of its cost. Exact harmonicity also ties the interior error to the boundary misfit through the maximum principle when every segment carries Dirichlet data. Our sampled estimate of that misfit closely matched the interior error wherever we could check it (\cref{sec:results}), so it can serve as a reference-free accuracy estimate. The hybrid model is more general in geometry and dimension, but keeps an interior residual that must be balanced against the boundary terms and a harmonic dictionary designed in advance. The holomorphic construction is restricted to two-dimensional, simply connected domains, and \cref{tab:dictionary} to straight-edged junctions with piecewise-constant data. On the sampled problems with Neumann data, only 17 of 48 runs reach a boundary loss below $10^{-6}$. More broadly, our results suggest that when a governing equation admits a sufficiently strong analytical structure, incorporating it directly into the neural hypothesis space can be more effective than enforcing it through a residual penalty. Rigorous enclosures of the boundary misfit, inverse and parametric problems, where the network's differentiability matters, and extensions to other geometries, higher dimensions and equations are left for future work.

\clearpage

\subsection*{Acknowledgements}

This work was partially supported by the \textit{i-RASE} project, funded by the European Union under the Horizon Europe EIC Pathfinder Open programme (Grant Agreement No. 101130550).

\bibliography{references}
\bibliographystyle{plainnat}

\clearpage

\appendix

\section{Enriched Holomorphic Networks}
\label{app:holomorphic}

\subsection{Implementation}
\label{app:holomorphic_implementation}

The model is implemented in JAX with 64-bit arithmetic enabled, so complex quantities are held in \texttt{complex128}. The accuracies reported in \cref{sec:results} are beyond what single precision can represent. Every complex weight is stored as a pair of real arrays and recombined in the forward pass, because the loss is real-valued and therefore not holomorphic in the weights, so differentiating it produces Wirtinger derivatives that the real-parameter optimizers support only partially. Weights are drawn from a scaled complex He distribution, with the real and imaginary parts of each entry independent and the layer variances chosen to hold the second moment of the pre-activations constant through the stack, following the initialization derived for holomorphic networks by \citet{calafa_physics-informed_2024}.

Fractional powers and logarithms such as $\zeta^{\lambda}$ and $\operatorname{Log}\zeta$ are multi-valued, and a single-valued branch is discontinuous along a ray from $z_c$. If that ray crossed the domain, the model would lose continuity and harmonicity there. Each site therefore stores the direction pointing out of the domain (the outward normal inside a face or an arc, the exterior bisector at a corner), and the cut is oriented along it. On every domain used here the cut then stays outside the domain.

We check the harmonicity through nested real automatic differentiation. For three randomly initialized cores on each of 18 problems, among them every named benchmark, variant and problem of \cref{tab:before_after}, $|\nabla^2 u|$ at 2\,000 random interior points stays below $10^{-14}$ for the core and below $3\times10^{-11}$ for each enrichment term in isolation.

For a problem with \emph{Neumann} data on the entire boundary, the solution is defined only up to an additive constant. This null direction would leave the objective unchanged and make the corresponding Hessian singular. We fix the gauge by replacing $f(z)$ with $f(z)-f(z_{\mathrm{anchor}})$ for a fixed anchor point $z_{\mathrm{anchor}}$. This subtraction preserves holomorphy and does not change the normal derivative, while fixing the additive constant of the predicted solution.

\subsection{Architecture and training}
\label{app:holomorphic_training}

Unless stated otherwise, the core has depth 2 and width 16, one enrichment term is placed per site, and both boundary weights in \cref{eq:boundary-loss} are set to one. On Motz's problem, its variant, and the cracked beam, the site instead carries the first four exponents of the \emph{Dirichlet--Neumann} ladder, $\tfrac12,\tfrac32,\tfrac52,\tfrac72$, pinned at these values; only the leading coefficient $D_0$ is compared with the literature, and we make no claim about the other three. Training is full-batch and proceeds in two stages: Adam at learning rate $10^{-2}$ with a cosine decay to zero, for $4\times10^{4}$ steps at the smallest of the four budgets of \cref{tab:all_models} and $1.5$, $2$ and $3$ times as many at the others, followed by $6 \times 10^3$ L-BFGS iterations with a history of 100 at every budget. The second stage is what decreases the loss from the level Adam reaches down to machine precision. Freezing an exponent during a stage is done by partitioning the parameter tree, so a frozen direction never enters the curvature memory of L-BFGS. Each run trains from three independent restarts and keeps the best, scored on a fresh boundary sample and never on a reference solution.

At the smallest budget, each segment is sampled at 1024 points per run, a number that scales with the Adam steps at the larger budgets. The runs of \cref{app:holomorphic_before_after} use 512. On the named benchmarks, their variants, and the problems of \cref{tab:before_after}, a further 128 points are drawn at each end of every segment, log-uniformly between $10^{-4}$ and $10^{-1}$ of its parameter range from the end, to resolve the $r^{\lambda}$ modes. The suite runs of \cref{fig:pairs} use the uniform points only. The exponents are trained on the two sectors, the L-shape, the slit disk, the pure switch mode, and the variants of the L-shape, the pure switch mode, and the $3\pi/2$ sector, and pinned everywhere else. The comparisons of \cref{app:lambda} and one row of \cref{tab:ablation} reverse this choice. Trained exponents share the network's learning rate and cosine schedule and enter the L-BFGS stage jointly with it. The core and optimizer settings and the number of uniform points were chosen by the ablation of \cref{tab:ablation}, on a tuning suite drawn with a seed disjoint from every evaluation suite, and the per-problem enrichment settings above (ladder depth, trained or pinned exponents, near-site points) were set by hand.

\subsection{Evaluation protocol}
\label{app:holomorphic_evaluation}

Problems with a closed-form solution are scored by relative $L^2$ error on interior points. Motz's problem and the cracked beam are scored by the relative error of the leading coefficient $D_0$ against its published value, and by relative $L^2$ error against a series reference: the expansion $\sum_{i=0}^{59} D_i\,r^{i+1/2}\cos\big((i+\tfrac12)\theta\big)$ about the singular point~\citep{li_singularities_2000}, which satisfies the two bottom edges exactly and is fitted by least squares on the other three, with a boundary residual of about $10^{-14}$. It reproduces the published $D_0$ of \citet{lu_highly_2004} to $10^{-15}$. Since Laplace's equation is linear, problems posed with large boundary data are solved at unit scale and the recovered coefficient rescaled afterwards, which avoids spending the capacity of the network on reaching the magnitude of the data.

The sampled maximum-principle estimate $\hat M$ of \cref{sec:holomorphic} draws $N$ uniform points on every \emph{Dirichlet} segment, discards those within $\rho=10^{-3}$ of a compiled site, and returns the largest remaining $|u_\theta-g|$, with $N=5\,000$ on the named benchmarks and the problems of \cref{app:holomorphic_before_after}, and $N=10^{4}$ on the suite of \cref{fig:pairs}. It omits the misfit between samples, inside the excluded discs and from rounding. We compare it with the interior maximum error over $1.3$--$2.6\times10^{5}$ points per model (uniform, graded geometrically down to distance $10^{-10}$ from every vertex, and refined around the largest errors), against the closed-form solution. On the square, that solution is computed through the conformal map to a half-plane. An independent finite-element solution, with fourth-order Lagrange elements~\citep{Gustafsson2020} on meshes graded to $10^{-14}$ at every vertex, agrees with it to $1.1\times10^{-8}$ away from the jumps and $6.5\times10^{-14}$ near them.

\paragraph{Continuous data.} On the two sectors, the L-shape and the slit disk, the interior maximum error of the enriched model lies within $2\times10^{-4}$ relative to $\hat M$ on all 45 models with $\hat M>4\times10^{-15}$ (41 of the 48 runs behind \cref{tab:all_models} and the four of \cref{tab:before_after}). In the other seven runs, $\hat M$ lies between $3\times10^{-16}$ and $3\times10^{-15}$, and the floating-point interior error exceeds it by up to a third. On the slit, $\hat M$ samples the upper face only, and sampling both faces leaves it unchanged on all 26 slit models we evaluated. The estimate fails when the misfit concentrates inside the excluded discs: with an exponent frozen $10\%$ away from its analytic value (\cref{app:lambda}), the interior error exceeds $\hat M$ by up to $37\%$, within $\rho$ of the site.

\paragraph{Data jumps.} The enriched Dirichlet-jump runs of \cref{fig:pairs} use the angle term $\mathrm{i}\,b\operatorname{Log}\zeta$ with real $b$, so the model stays bounded at the jumps. On the square, the interior maximum error exceeds $\hat M$ by $3$--$11\%$ at distances of at least $10^{-5}$ from the jumps and by at most $12\%$ anywhere, because the largest boundary misfit lies inside the excluded discs. Counting the misfit there closes the gap to about $1\%$. On the disk, the interior maximum error is within $1.8\%$ of $\hat M$ at distances of at least $10^{-6}$ from the jumps. Closer in, the ratio grows to about $10^{3}$ at $10^{-10}$, but this is a floating-point effect, not a model error: a jump point $e^{2\pi\mathrm{i}t}$ is generally not representable exactly, so the singular term sits up to $5\times10^{-16}$ from the true jump, and this offset accounts for the excess to within $8\%$.

The plain network cannot represent a jump, so its error is at least half the largest jump, and its $\hat M$ underestimates the error on 23 of its 24 runs. The two disks of \cref{app:holomorphic_before_after} were trained with a complex coefficient on $\operatorname{Log}\zeta$, so their models are unbounded at the jumps and \cref{eq:estimate} does not apply. Their error exceeds $\hat M$ by at most $13\%$ at distances of at least $10^{-5}$ from the jumps, and by up to a factor of $4.75$ at $10^{-10}$.

\subsection{Scope of the enrichment dictionary}
\label{app:dictionary}

\begin{figure}[ht]
    \centering
    \includegraphics[width=0.72\linewidth]{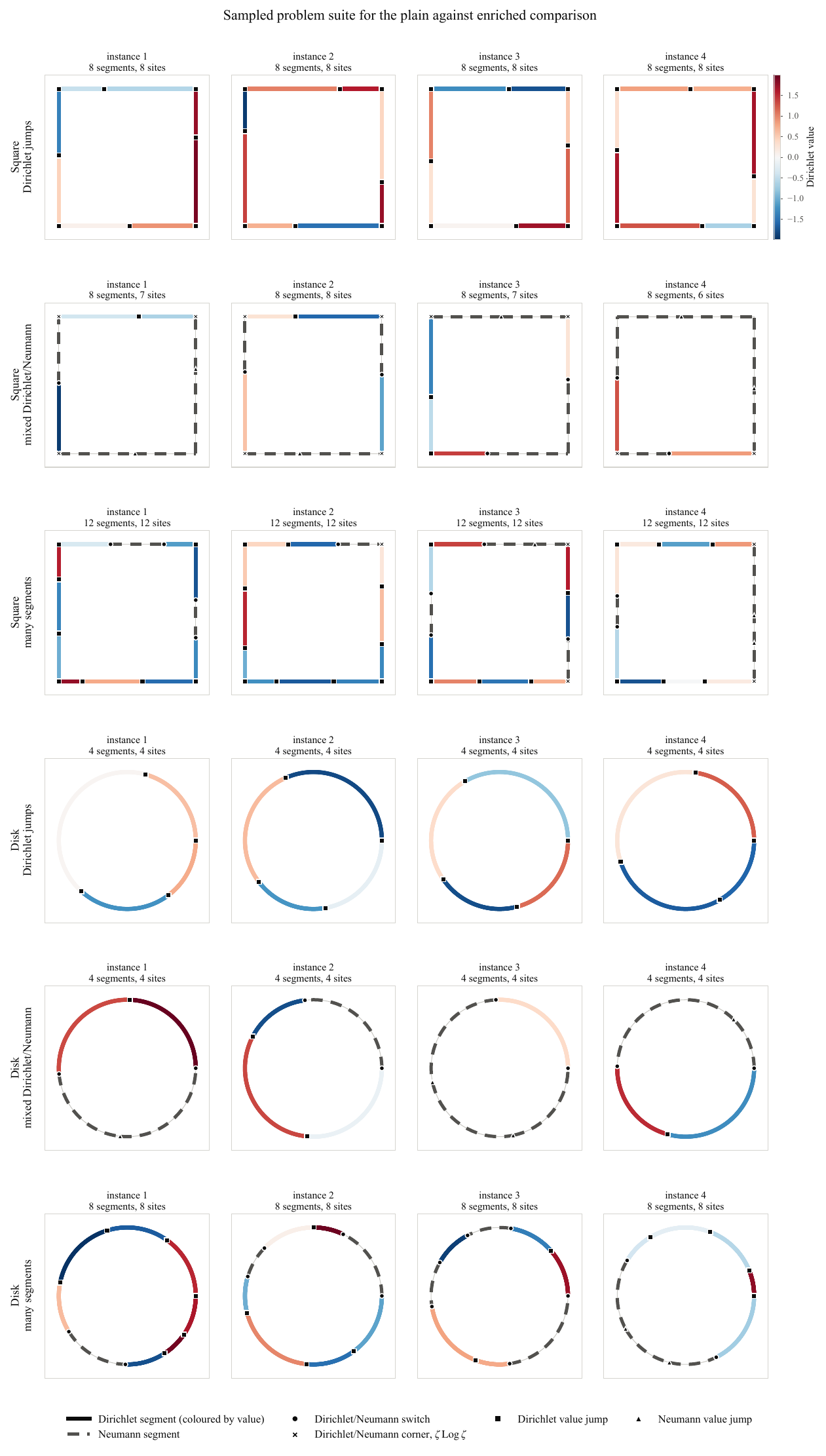}
    \caption{The 24 sampled instances used for the comparison of \cref{fig:pairs}, four per domain and family. \emph{Dirichlet} segments are coloured by their prescribed value on a scale shared by every panel; \emph{Neumann} segments carry a flux, which is not on the same scale as $u$, and are drawn dashed. Markers are the singular sites the compiler emits, with the shape giving the row of \cref{tab:dictionary} and crosses marking the \emph{Dirichlet--Neumann} corners with nonzero flux, which receive $\zeta\operatorname{Log}\zeta$.}
    \label{fig:suite_eval}
\end{figure}

\begin{figure}[ht]
    \centering
    \includegraphics[width=0.85\linewidth]{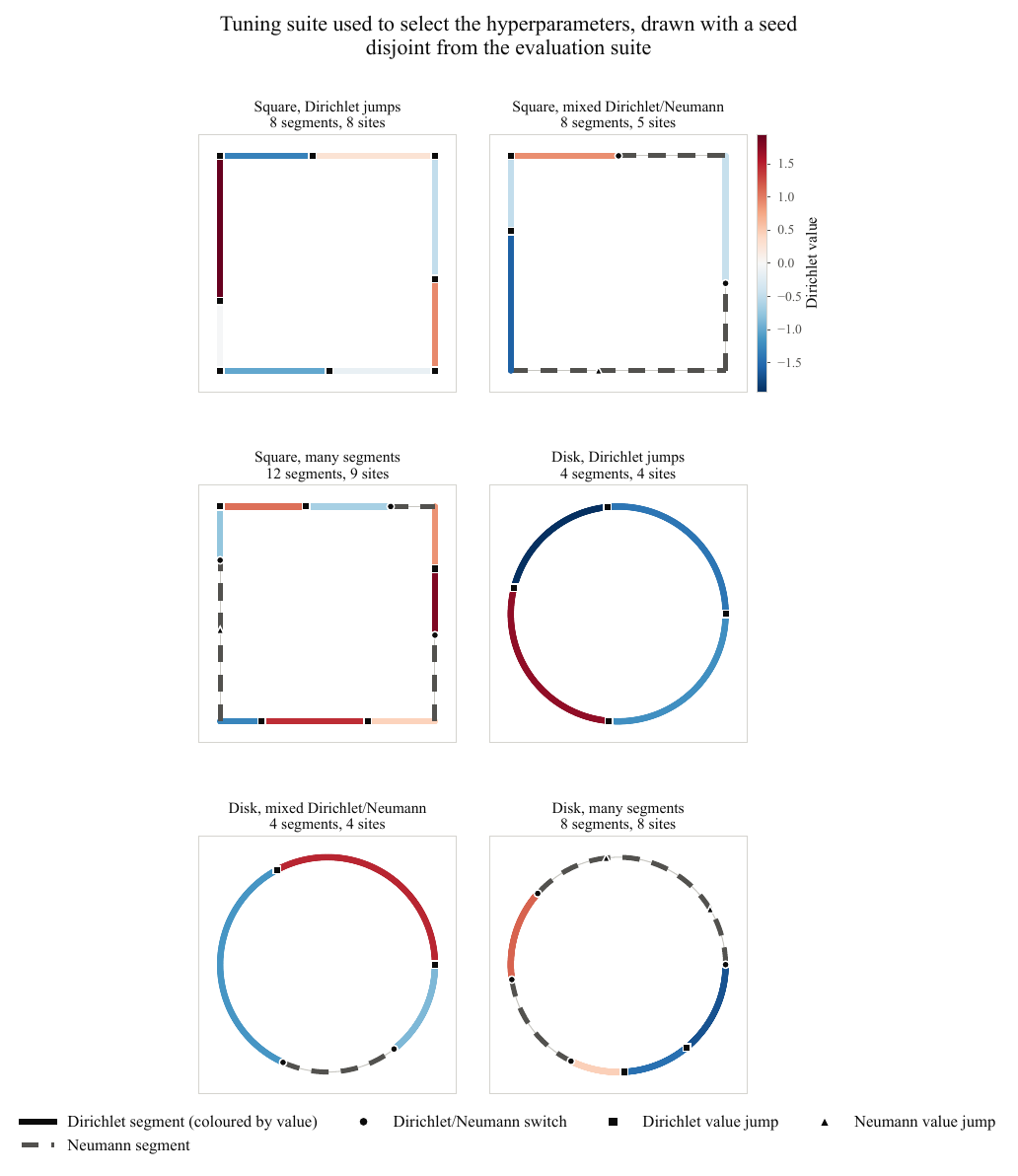}
    \caption{The tuning suite used for the ablation of \cref{tab:ablation}, drawn from the same generator with a different seed and disjoint from the evaluation suite of \cref{fig:suite_eval}, so no hyperparameter of \cref{tab:ablation} was selected on a problem it is reported on. Its markers are the sites of the runs of \cref{tab:ablation}, which do not include the $\zeta\operatorname{Log}\zeta$ term at its five \emph{Dirichlet--Neumann} corners with nonzero flux.}
    \label{fig:suite_tuning}
\end{figure}

The compiler walks the boundary, a sequence of straight segments and circular arcs with their conditions and data, and gives each junction the terms of every row of \cref{tab:dictionary} it matches. It refuses what it cannot place: a crack whose faces do not lie along a coordinate axis, a crack mouth that needs terms, and a branch cut that would re-enter a non-convex domain. On the named benchmarks and their variants, the compiled sites agree to the last bit with sites written by hand. The $\zeta\operatorname{Log}\zeta$ row applies at 16 \emph{Dirichlet--Neumann} corners in 7 of the 8 square instances with \emph{Neumann} data in the evaluation suite (\cref{fig:suite_eval}) and at 5 corners of the tuning suite (\cref{fig:suite_tuning}). The runs of \cref{fig:pairs} include this term, those of \cref{tab:ablation} do not.

\Cref{tab:dictionary} follows the local expansion of the solution at a junction of straight edges with piecewise-constant data~\citep{kondratev_boundary_1967,grisvard_elliptic_1985}. A \emph{Neumann--Neumann} junction has the exponents of a \emph{Dirichlet--Dirichlet} one, with cosine instead of sine modes, which a complex coefficient on $\zeta^{\lambda}$ also covers; this includes re-entrant \emph{Neumann--Neumann} corners, although none of our problems has one. Beyond this class the dictionary is incomplete. Non-constant data change the forcing and can resonate at higher degrees, adding $\zeta^p\operatorname{Log}\zeta$. \emph{Dirichlet} data $x^2+y^2$ at a square corner, for example, give an $r^2\log r$ term~\citep{lehman_developments_1959,wigley_asymptotic_1964}. The non-constant data in our problems are traces of harmonic functions that are analytic at the junctions, so no such term arises. On curved edges a \emph{Neumann} jump also carries $\zeta^p\operatorname{Log}\zeta$ for every $p\ge2$, and junctions are enriched with the exponents of the tangent wedge only, and for \emph{Dirichlet} jumps on the disk this is exact, since the solution is a constant plus angle terms.
\section{Problem Suites}
\label{app:holomorphic_suites}

\begin{figure}[ht]
    \centering
    \includegraphics[width=\linewidth]{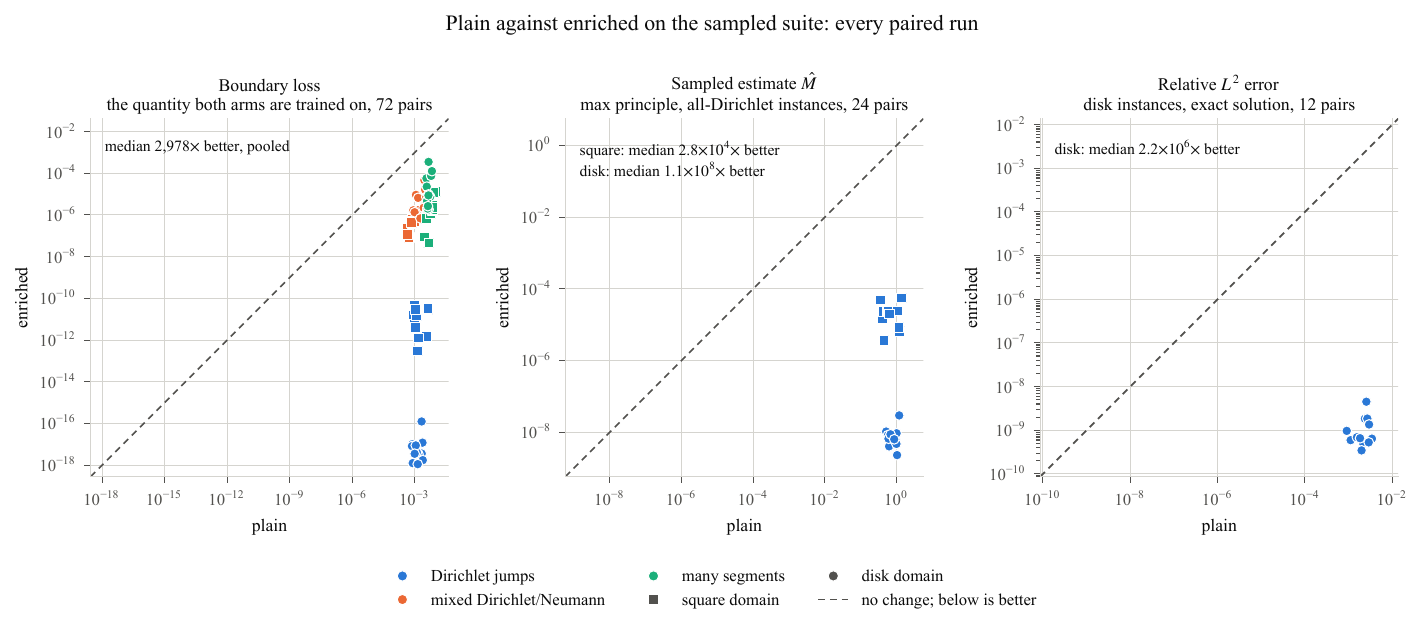}
    \caption{Plain against enriched over the sampled suite, one point per paired run: the same instance and the same seed trained twice, with the singular terms and without, plain on the horizontal axis and enriched on the vertical. The diagonal is where enrichment changed nothing, so everything below it is a run that improved. The panels cover 72, 24 and 12 pairs: the boundary loss of every run; the sampled maximum-principle estimate $\hat M$ of \cref{sec:holomorphic} on the 24 runs with \emph{Dirichlet} data on every segment; and the relative $L^2$ error on the 12 disk Dirichlet-jump runs. Colour is the family of the instance, and the marker follows the shape of its domain.}
    \label{fig:pairs}
\end{figure}

\Cref{tab:suite_families} splits \cref{fig:pairs} by family. The enriched model converges, with a boundary loss below $10^{-6}$, on all 24 Dirichlet-jump runs and on 17 of the 48 runs with \emph{Neumann} data.

A least-squares fit without a network shows what the compiled terms contribute. On the disk Dirichlet-jump instances, whose solutions are a constant plus angle terms of \cref{tab:dictionary}, a fit of the compiled terms alone reaches a relative $L^2$ error of $6.1\times10^{-14}$ or better, and on the other 20 instances it leaves a boundary loss of $4.4\times10^{-3}$ to $0.54$. With eight ladder terms per site and harmonic polynomials up to degree 48, the fit reaches a lower boundary loss than the enriched network on all 24 instances (median $4.3\times10^{-20}$ against $7.1\times10^{-7}$), in 0.7 to 2.3~s against a median of 952 to 6\,634~s per network run, although on four disk instances with \emph{Neumann} data it stops between $3.2\times10^{-7}$ and $1.2\times10^{-6}$. On the seven square instances with a \emph{Dirichlet--Neumann} corner with nonzero flux, the same fit without the $\zeta\operatorname{Log}\zeta$ term stops at $2.4\times10^{-7}$ to $3.2\times10^{-6}$, and with it reaches $3\times10^{-17}$ or below. Because a fit with more terms per site reaches a lower loss on every instance, the network's shortfall on the \emph{Neumann} families may come from its single term per site or from its optimization. We have not separated the two. These fits, as the least-squares fits of \cref{tab:all_models}, use 800 uniform boundary points per segment and 200 at each end, log-spaced from $10^{-1}$ down to $10^{-10}$ of the segment ($10^{-5}$ on the disk), with poles placed by AAA~\citep{nakatsukasa_aaa_2018} where used.

\begin{table}[ht]
  \caption{The sampled suite of \cref{fig:pairs} by family: four instances and three seeds per family, paired. The two ratios divide the plain run's value by the enriched run's: the boundary-loss ratio gives the median with the minimum and maximum over the 12 pairs, the $\hat M$ ratio the median. \emph{Enriched loss} is the median boundary loss of the enriched runs, and \emph{Conv.} counts those below $10^{-6}$. Both arms use the same steps, samples, and three restarts, with hyperparameters tuned for the enriched arm. \emph{Time} is the ratio of median wall-clock time per run, enriched over plain.}
  \label{tab:suite_families}
  \centering\footnotesize
  \setlength{\tabcolsep}{3pt}
  \begin{tabular}{lccccc}
    \toprule
    Family & Boundary-loss ratio & Enriched loss & Conv. & $\hat M$ ratio & Time\\
    \midrule
    Disk, D jumps & $3.0{\times}10^{14}$ [$1.7{\times}10^{13}$, $1.4{\times}10^{15}$] & $3.8{\times}10^{-18}$ & 12/12 & $1.1{\times}10^{8}$ & 1.11\\
    Square, D jumps & $1.1{\times}10^{8}$ [$2.0{\times}10^{7}$, $4.6{\times}10^{9}$] & $1.5{\times}10^{-11}$ & 12/12 & $2.8{\times}10^{4}$ & 1.30\\
    Disk, mixed D/N & 616 [61, $4.9{\times}10^{3}$] & $2.5{\times}10^{-6}$ & 3/12 & -- & 1.13\\
    Square, mixed D/N & $1.8{\times}10^{3}$ [$1.1{\times}10^{3}$, $6.7{\times}10^{3}$] & $4.9{\times}10^{-7}$ & 10/12 & -- & 1.38\\
    Disk, many segments & 498 [13, $2.5{\times}10^{3}$] & $9.1{\times}10^{-6}$ & 0/12 & -- & 1.37\\
    Square, many segments & $3.6{\times}10^{3}$ [693, $1.1{\times}10^{5}$] & $1.5{\times}10^{-6}$ & 4/12 & -- & 1.50\\
    \bottomrule
  \end{tabular}
\end{table}

\FloatBarrier

\section{Named Benchmarks}
\label{app:holomorphic_benchmarks}

\begin{figure}[!ht]
    \centering
    \includegraphics[width=0.95\linewidth]{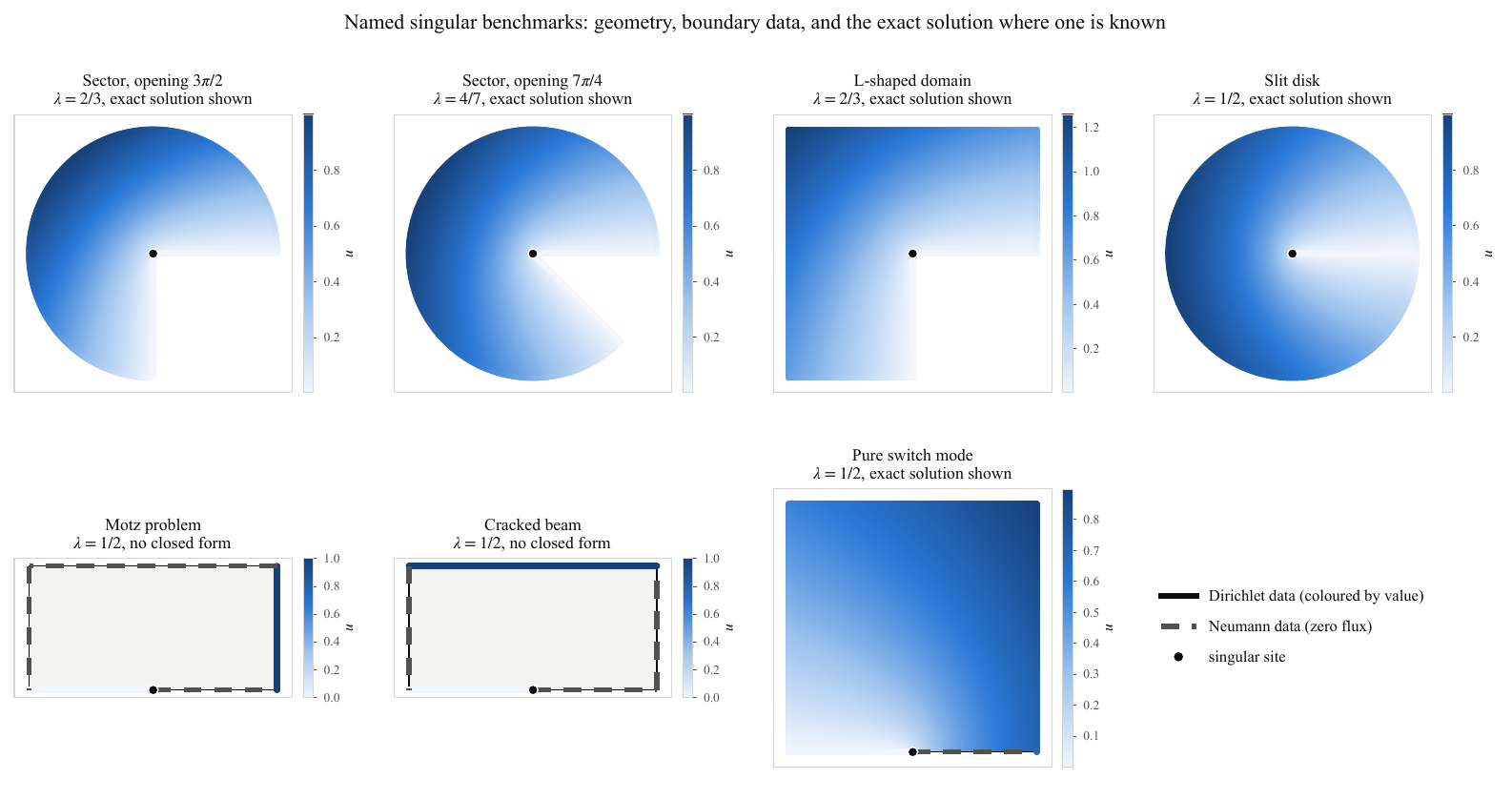}
    \caption{The six benchmarks of \cref{tab:benchmarks}, together with the pure switch mode of \cref{tab:before_after}, with the exact solution filling the interior where one exists.}
    \label{fig:benchmarks_overview}
\end{figure}

The six named problems of \cref{tab:all_models} and its four variants are listed in \cref{tab:benchmarks}; the named problems are drawn in \cref{fig:benchmarks_overview}. Each variant keeps the geometry, boundary types and singular site of its parent and adds a smooth harmonic function to its singular mode, so its data are the traces of a known solution; for Motz with a pole, the singular part is the switch mode $\operatorname{Re}z^{1/2}$ alone.

\begin{table}[!ht]
  \caption{The named singular benchmarks and the four variants of \cref{tab:all_models}. $\lambda$ is the leading exponent that \cref{tab:dictionary} assigns to the single site, which the compiler finds. \emph{Terms} is the number of ladder terms at the site, with exponents trained from their value in \cref{tab:dictionary} or pinned there. For the first four problems the exact solution is the single enrichment term itself. Reference values for the two coefficient problems are from \citet{lu_highly_2004}, who obtain them to 17 significant digits with a collocation Trefftz method. Motz's problem is due to \citet{motz_treatment_1947}; for the cracked beam we use the domain and boundary conditions of \citet{lu_highly_2004}, so that the published coefficient applies.}
  \label{tab:benchmarks}
  \centering
  \small
  \setlength{\tabcolsep}{4pt}
  \begin{tabular}{lllll}
    \toprule
    Problem & Domain & $\lambda$ & Terms & Reference \\
    \midrule
    Sector $3\pi/2$ & circular sector, opening $3\pi/2$ & $2/3$ & 1, trained & exact \\
    Sector $7\pi/4$ & circular sector, opening $7\pi/4$ & $4/7$ & 1, trained & exact \\
    L-shape & square minus a quadrant & $2/3$ & 1, trained & exact \\
    Slit disk & disk cut from centre to rim & $1/2$ & 1, trained & exact \\
    Motz & rectangle, D/N switch mid-edge & $1/2$ & 4, pinned & $D_0 = 401.16245374523$ \\
    Cracked beam & same rectangle, conditions swapped & $1/2$ & 4, pinned & $D_0 = 540.56512271363$ \\
    \midrule
    L-shape $+\operatorname{Re}e^{z}$ & as L-shape & $2/3$ & 1, trained & exact \\
    Pure mode $+\operatorname{Re}e^{z}$ & unit square, D/N switch mid-edge & $1/2$ & 1, trained & exact \\
    Sector $3\pi/2$ + pole & as sector $3\pi/2$, pole at $1.1e^{\mathrm{i}\pi/4}$ & $2/3$ & 1, trained & exact \\
    Motz + pole & as Motz, pole at $0.3-0.25\mathrm{i}$ & $1/2$ & 4, pinned & exact \\
    \bottomrule
  \end{tabular}
\end{table}

\Cref{tab:ladder_seeds} gives every training budget of the holomorphic networks on the named problems. Each budget multiplies the Adam steps and the boundary points per segment, from $\times1$ to $\times3$, with $6\times10^{3}$ L-BFGS iterations throughout; \cref{tab:all_models} reports, for each method and metric, the median over three seeds at the budget where that median is lowest, a choice made after the fact.

The three other models of \cref{tab:all_models} are implemented in JAX in double precision and scored on the same interior points as the holomorphic networks. The MLP PINN has one hidden layer of 128 tanh units (513 parameters). It minimises the mean squared Laplacian at 5\,000 fixed interior points, a fifth of them within $0.1$ of the singular site, plus the mean squared error of the boundary data at 400 boundary points, with weight 10 on the \emph{Dirichlet} data on the two sectors and 100 elsewhere, by $4\times10^{4}$ Adam steps at learning rate $10^{-3}$ followed by $6\times10^{3}$ L-BFGS iterations; each budget multiplies the Adam steps and the boundary points. The harmonic fit is a least-squares fit of the boundary data, with \emph{Dirichlet} weight 10, in harmonic polynomials of degree up to 20 on the sectors and 40 elsewhere plus 40 Poisson kernels attached to boundary points (81 or 121 coefficients). The hybrid model adds such a fit, made with its own \emph{Dirichlet} weight and then frozen, to an MLP PINN trained on the remaining residual (594 or 634 parameters); the holomorphic networks have 642 to 654 real parameters. For these models and the plain network, $D_0$ is estimated by a least-squares fit of the first six terms of the series of \cref{app:holomorphic_evaluation} to the model at $0.02\le r\le0.4$; the enriched network's $D_0$ is its own leading coefficient.

Across the budgets, the MLP PINN improves by at most a factor of $1.2$ and the harmonic fit changes by at most $4\%$, its error being truncation rather than optimization. The hybrid model improves on that fit by a factor of $1.05$ to $2.0$ on the closed-form problems but is $2.9$ and $2.0$ times less accurate on Motz's problem and the cracked beam, where its frozen fit, made with \emph{Dirichlet} weight 100, already has errors of $3.6\times10^{-2}$ and $3.0\times10^{-2}$ against $1.7\times10^{-2}$ for the harmonic fit. The plain holomorphic network improves by a factor of $1.7$ to $3$ on the closed-form problems and $5.6$ and $3.3$ on Motz's problem and the cracked beam, where its estimated $D_0$ error falls from about $6\times10^{-3}$ and $5\times10^{-3}$ to $4.2\times10^{-4}$ and $5.9\times10^{-4}$. Only the enriched network gains orders of magnitude on the closed-form problems, and its outcomes are bimodal across seeds: at the largest budget the L-shape ranges from $2.3\times10^{-16}$ to $1.5\times10^{-11}$ and the slit disk from $1.7\times10^{-16}$ to $7.9\times10^{-12}$, so the rounding-level median of the L-shape rests on two of three seeds. The cracked-beam $D_0$ error is not monotone in budget; its entry in \cref{tab:all_models} is the median at the smallest budget.

\begin{table}[ht]
  \caption{The named benchmarks at every training budget of the holomorphic networks: median over three seeds, with the range over seeds below it for the enriched network. Each budget multiplies the Adam steps and the boundary points per segment ($4\times10^{4}$ and 1024 at $\times1$), with $6\times10^{3}$ L-BFGS iterations throughout; both networks use the configuration tuned for the enriched one. Motz's problem and the cracked beam are scored against the series reference of \cref{app:holomorphic_evaluation}.}
  \label{tab:ladder_seeds}
  \centering\scriptsize
  \setlength{\tabcolsep}{2pt}
  \begin{tabular}{@{}llcccc@{}}
    \toprule
    Problem & Model & $\times1$ & $\times1.5$ & $\times2$ & $\times3$\\
    \midrule
    \multicolumn{6}{@{}l}{\emph{Relative $L^2$ error}}\\
    Sector $3\pi/2$ & enriched & $8.3{\times}10^{-11}$ & $4.4{\times}10^{-11}$ & $2.1{\times}10^{-11}$ & $3.3{\times}10^{-12}$\\
     & range & $7.2{\times}10^{-11}$--$1.1{\times}10^{-10}$ & $3.3{\times}10^{-13}$--$5.9{\times}10^{-11}$ & $1.8{\times}10^{-11}$--$2.8{\times}10^{-11}$ & $2.9{\times}10^{-12}$--$4.1{\times}10^{-12}$\\
     & plain & $1.7{\times}10^{-3}$ & $1.2{\times}10^{-3}$ & $1.2{\times}10^{-3}$ & $7.7{\times}10^{-4}$\\
    \addlinespace
    Sector $7\pi/4$ & enriched & $6.0{\times}10^{-11}$ & $7.1{\times}10^{-11}$ & $1.9{\times}10^{-11}$ & $1.4{\times}10^{-16}$\\
     & range & $4.4{\times}10^{-11}$--$1.3{\times}10^{-10}$ & $2.0{\times}10^{-11}$--$8.9{\times}10^{-11}$ & $2.5{\times}10^{-12}$--$2.7{\times}10^{-11}$ & $1.4{\times}10^{-16}$--$9.0{\times}10^{-16}$\\
     & plain & $3.1{\times}10^{-2}$ & $2.3{\times}10^{-2}$ & $2.1{\times}10^{-2}$ & $1.9{\times}10^{-2}$\\
    \addlinespace
    L-shape & enriched & $4.1{\times}10^{-11}$ & $9.8{\times}10^{-12}$ & $9.9{\times}10^{-12}$ & $2.9{\times}10^{-16}$\\
     & range & $3.3{\times}10^{-11}$--$7.9{\times}10^{-11}$ & $3.6{\times}10^{-12}$--$1.0{\times}10^{-11}$ & $2.6{\times}10^{-16}$--$3.1{\times}10^{-11}$ & $2.3{\times}10^{-16}$--$1.5{\times}10^{-11}$\\
     & plain & $1.5{\times}10^{-3}$ & $1.4{\times}10^{-3}$ & $1.0{\times}10^{-3}$ & $7.5{\times}10^{-4}$\\
    \addlinespace
    Slit disk & enriched & $8.4{\times}10^{-11}$ & $9.3{\times}10^{-12}$ & $1.1{\times}10^{-11}$ & $2.8{\times}10^{-12}$\\
     & range & $3.2{\times}10^{-11}$--$1.0{\times}10^{-10}$ & $3.9{\times}10^{-12}$--$2.3{\times}10^{-11}$ & $5.2{\times}10^{-12}$--$1.8{\times}10^{-11}$ & $1.7{\times}10^{-16}$--$7.9{\times}10^{-12}$\\
     & plain & $9.8{\times}10^{-1}$ & $9.7{\times}10^{-1}$ & $3.4{\times}10^{-1}$ & $3.2{\times}10^{-1}$\\
    \addlinespace
    Motz & enriched & $1.9{\times}10^{-7}$ & $1.2{\times}10^{-7}$ & $2.5{\times}10^{-8}$ & $2.7{\times}10^{-8}$\\
     & range & $8.7{\times}10^{-8}$--$2.6{\times}10^{-7}$ & $8.9{\times}10^{-8}$--$1.5{\times}10^{-7}$ & $2.0{\times}10^{-8}$--$5.8{\times}10^{-8}$ & $2.4{\times}10^{-8}$--$6.1{\times}10^{-8}$\\
     & plain & $3.1{\times}10^{-3}$ & $1.7{\times}10^{-3}$ & $1.0{\times}10^{-3}$ & $5.4{\times}10^{-4}$\\
    \addlinespace
    Cracked beam & enriched & $3.3{\times}10^{-7}$ & $2.8{\times}10^{-7}$ & $1.8{\times}10^{-7}$ & $8.6{\times}10^{-8}$\\
     & range & $3.1{\times}10^{-7}$--$7.3{\times}10^{-7}$ & $2.1{\times}10^{-7}$--$3.7{\times}10^{-7}$ & $1.7{\times}10^{-7}$--$2.0{\times}10^{-7}$ & $8.4{\times}10^{-8}$--$2.7{\times}10^{-7}$\\
     & plain & $2.1{\times}10^{-3}$ & $8.2{\times}10^{-4}$ & $1.2{\times}10^{-3}$ & $6.4{\times}10^{-4}$\\
    \midrule
    \multicolumn{6}{@{}l}{\emph{Relative error of $D_0$}}\\
    Motz & enriched & $4.5{\times}10^{-6}$ & $5.8{\times}10^{-6}$ & $2.3{\times}10^{-6}$ & $8.2{\times}10^{-7}$\\
     & range & $3.8{\times}10^{-6}$--$5.2{\times}10^{-6}$ & $3.8{\times}10^{-6}$--$7.5{\times}10^{-6}$ & $8.0{\times}10^{-7}$--$1.2{\times}10^{-5}$ & $6.8{\times}10^{-7}$--$1.5{\times}10^{-6}$\\
    \addlinespace
    Cracked beam & enriched & $2.0{\times}10^{-6}$ & $1.1{\times}10^{-5}$ & $5.9{\times}10^{-6}$ & $2.7{\times}10^{-6}$\\
     & range & $3.8{\times}10^{-7}$--$6.2{\times}10^{-6}$ & $5.2{\times}10^{-6}$--$3.1{\times}10^{-5}$ & $5.4{\times}10^{-6}$--$8.4{\times}10^{-6}$ & $2.4{\times}10^{-6}$--$5.0{\times}10^{-5}$\\
    \bottomrule
  \end{tabular}
\end{table}

\FloatBarrier

\section{Enrichment Against a Plain Holomorphic Network}
\label{app:holomorphic_before_after}

\begin{figure}[!ht]
    \centering
    \includegraphics[width=0.86\linewidth]{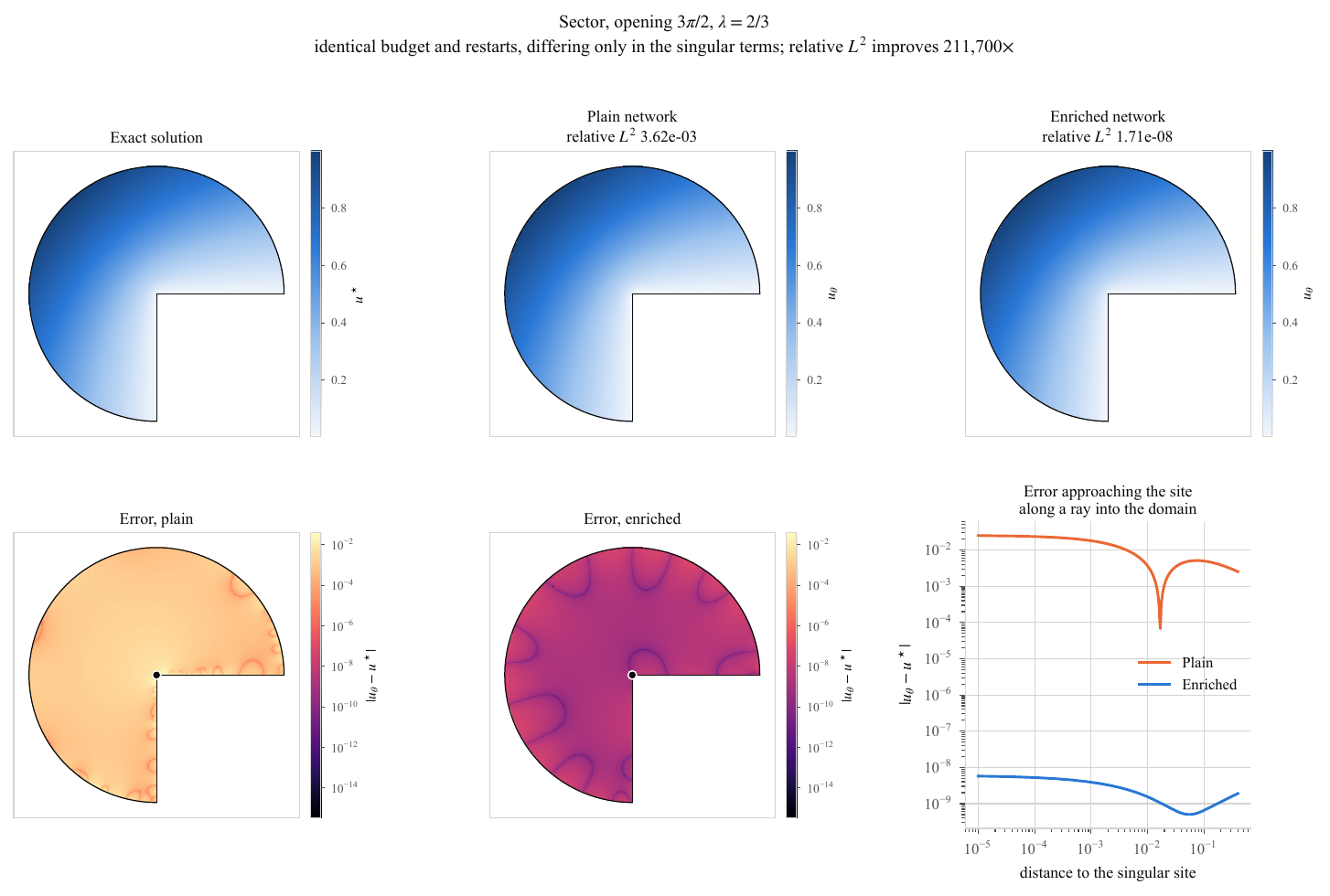}
    \caption{The two models on the sector of opening $3\pi/2$. Top row: the exact solution and the two predictions, on a common scale. Bottom row: the corresponding errors, on a second common scale, and the error along a ray directed into the singular site.}
    \label{fig:before_after}
\end{figure}

\Cref{tab:before_after} reports the seven problems outside the sampled suite that have both a closed-form solution and at least one singular site. In each case, the two models are trained at the same budget and with the same three restarts, so that the singular terms are the only difference between them. All seven solutions lie in the span of the singular terms (a single power term, or one angle term per jump on the disks), so these pairs measure what the singular terms add to a network rather than the network's own approximation power: a least-squares fit of the singular terms alone is $10^{6}$ to $10^{8}$ times more accurate than the enriched network here (last column of \cref{tab:before_after}). \Cref{fig:pairs} reports aggregate statistics over the sampled suite, whereas the figures below show the spatial distribution of the error. For the plain model, the error is spread over the entire domain and is largest in the vicinity of the sites; for the enriched model, it is smooth away from the sites and smaller by at least four orders of magnitude.

The runs in this section use a smaller budget than the ladder behind \cref{tab:all_models} and a single seed: $8 \times 10^{3}$ Adam steps, $2 \times 10^{3}$ L-BFGS iterations and 512 boundary points per segment, against $4 \times 10^{4}$, $6 \times 10^{3}$ and 1024 at the smallest point of that ladder and three times as many Adam steps and points, with the same L-BFGS iterations, at the largest. The absolute errors reported below are correspondingly larger. On the four problems common to both, the enriched model is here between $1.6 \times 10^{2}$ and $1.7 \times 10^{3}$ times less accurate than at the smallest ladder budget; the slit disk attains $1.3\times10^{-8}$ here and $1.7\times10^{-16}$ on its best seed at the largest ladder budget, against the three-seed median of $2.8\times10^{-12}$ reported in \cref{tab:all_models}. Since the two models receive the same budget, the comparison is unaffected, but the errors in \cref{tab:before_after} are not the accuracy attainable by the enriched model. On those four problems, the plain model was also run on the ladder (\cref{tab:ladder_seeds}), where only the enriched model improves by orders of magnitude; the pure switch mode and the two disks are not part of the named benchmark set and were run at this budget only.

\begin{table}[!ht]
  \caption{The paired runs behind the figures of this section: one seed at the reduced budget given above, three restarts each; the two networks differ only in the singular terms. Relative $L^2$ error against the closed-form solution. Each of these solutions lies in the span of the singular terms; the last column (\emph{Enr.\ LS}) fits those terms alone by least squares, without a network, in about a second. ${}^{*}$On the disks the fit's boundary points approach each jump only to $10^{-5}$ of the arc, against $10^{-10}$ on the other problems.}
  \label{tab:before_after}
  \centering
  \begin{tabular}{lrrrrr}
    \toprule
    Problem & Sites & Plain & Enriched & Factor & Enr.\ LS \\
    \midrule
    Sector, opening $3\pi/2$ & 1 & $3.6 \times 10^{-3}$ & $1.7 \times 10^{-8}$ & $2.1 \times 10^{5}$ & $5.5 \times 10^{-16}$ \\
    Sector, opening $7\pi/4$ & 1 & $5.4 \times 10^{-2}$ & $9.6 \times 10^{-9}$ & $5.6 \times 10^{6}$ & $9.9 \times 10^{-16}$ \\
    L-shaped domain & 1 & $4.0 \times 10^{-3}$ & $7.1 \times 10^{-8}$ & $5.7 \times 10^{4}$ & $5.3 \times 10^{-16}$ \\
    Pure switch mode & 1 & $7.6 \times 10^{-3}$ & $1.7 \times 10^{-7}$ & $4.5 \times 10^{4}$ & $2.0 \times 10^{-15}$ \\
    Slit disk & 1 & $1.2$ & $1.3 \times 10^{-8}$ & -- & $2.0 \times 10^{-16}$ \\
    Disk, four \emph{Dirichlet} arcs & 4 & $1.1 \times 10^{-2}$ & $6.5 \times 10^{-8}$ & $1.7 \times 10^{5}$ & $1.5 \times 10^{-14}$\rlap{${}^{*}$} \\
    Disk, eight alternating arcs & 8 & $3.7 \times 10^{-2}$ & $3.1 \times 10^{-7}$ & $1.2 \times 10^{5}$ & $2.6 \times 10^{-13}$\rlap{${}^{*}$} \\
    \bottomrule
  \end{tabular}
\end{table}

\begin{figure}[!ht]
    \centering
    \includegraphics[width=0.86\linewidth]{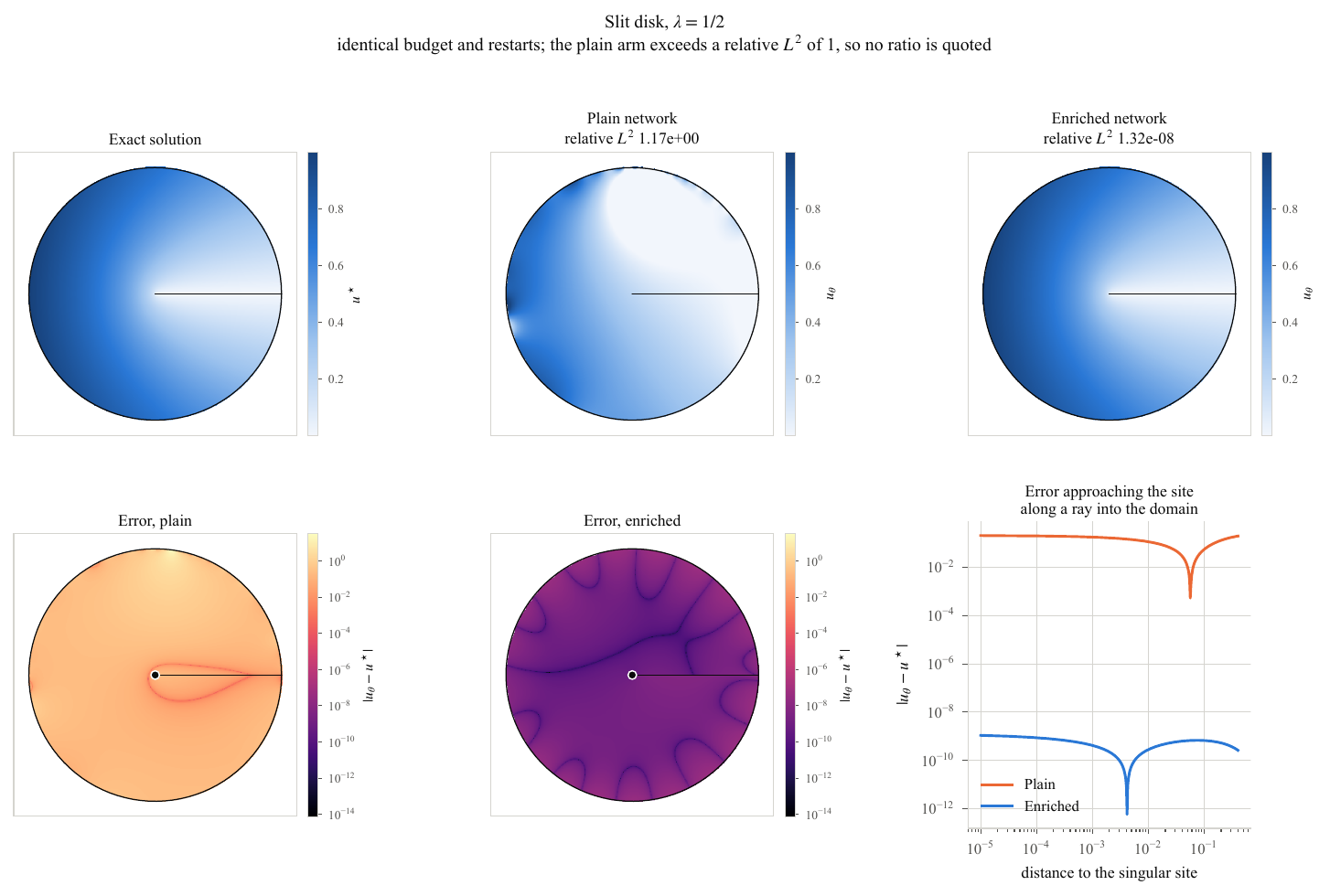}
    \caption{The slit disk. The plain model fails over the whole domain and not only near the tip, with a relative $L^2$ error of 1.17 against $1.3 \times 10^{-8}$ for the enriched model. The enrichment term supplies the branch that an entire core cannot.}
    \label{fig:before_after_slit}
\end{figure}

No improvement factor is given for the slit disk in \cref{fig:before_after_slit}. The relative $L^2$ error of the plain model exceeds unity, i.e., it is larger than that of the zero function. The two faces of the slit consist of the same points with opposite normals; the solution vanishes on both, but its derivative normal to the slit changes sign across it, whereas the core of the model is entire and therefore smooth across the slit. The plain network, which has only that core, still has a relative $L^2$ error of $0.32$ at the largest budget of \cref{tab:ladder_seeds}. Only the branch-power term, whose cut is placed along the slit itself, can represent the solution, so the gap on the slit is one of representation, not of training budget, and a ratio against the plain model is not meaningful.

\begin{figure}[!ht]
    \centering
    \includegraphics[width=0.86\linewidth]{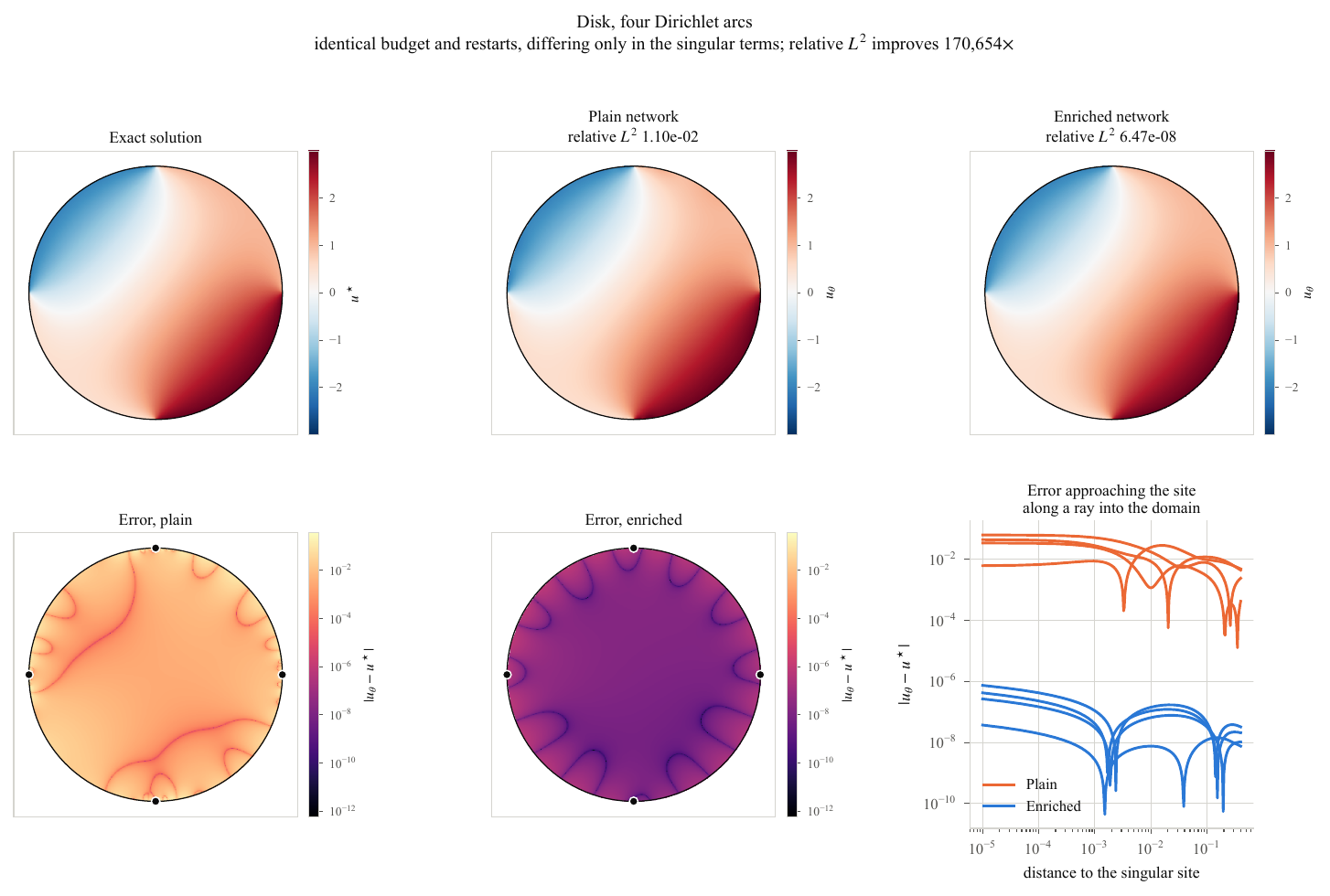}
    \caption{A disk with four \emph{Dirichlet} arcs, giving four value jumps and four compiled sites. The last panel shows one curve per site for each model. The exact solution changes sign here, unlike in the single-site problems, and the top row is therefore drawn on a diverging scale.}
    \label{fig:before_after_arcs}
\end{figure}

\begin{figure}[!ht]
    \centering
    \includegraphics[width=0.86\linewidth]{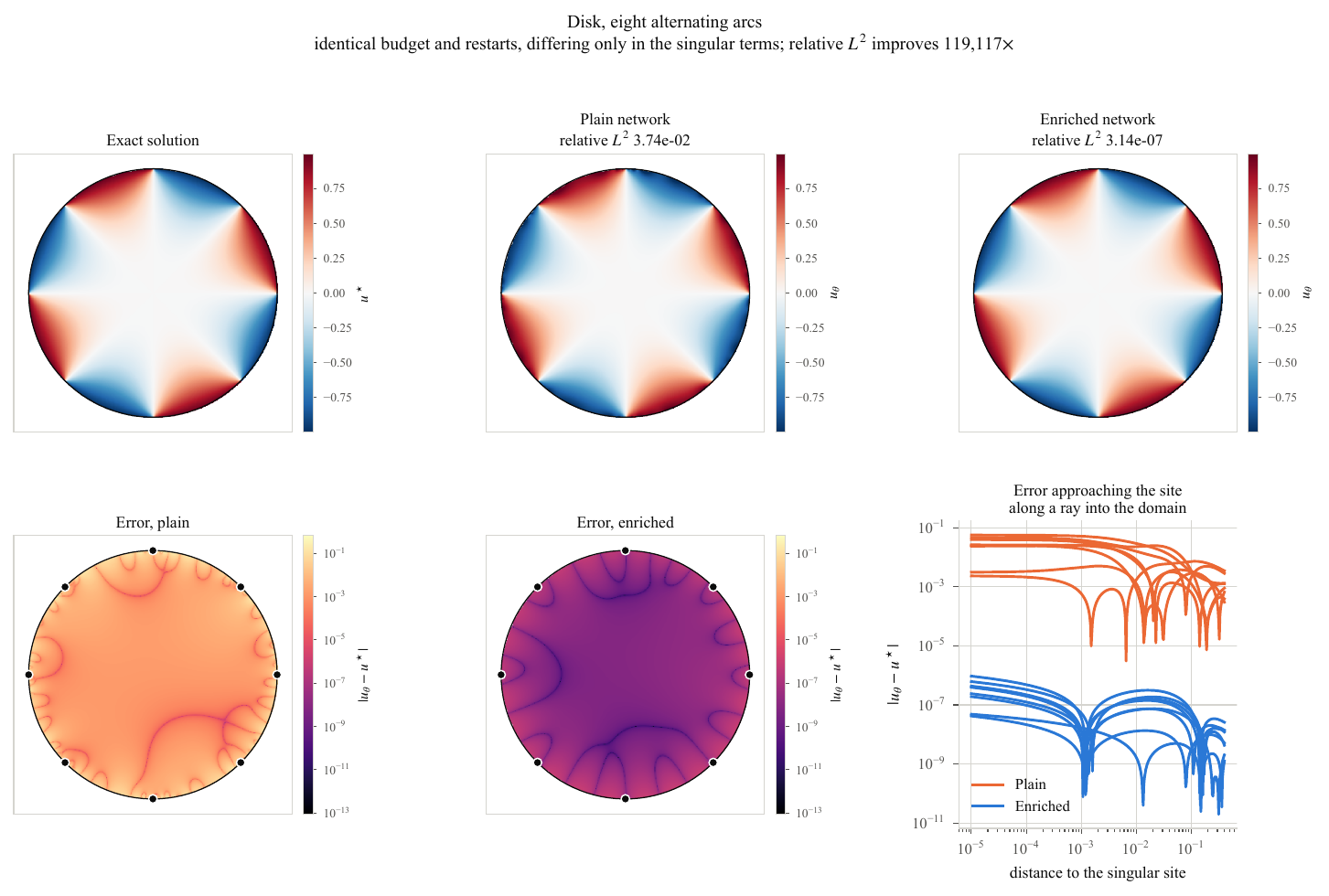}
    \caption{The same disk with eight alternating arcs, giving eight sites. The two models remain separated at every site, and the errors are comparable to those obtained with four sites.}
    \label{fig:before_after_multi}
\end{figure}

The last two rows of \cref{tab:before_after} concern domains with more than one singular site. All named benchmarks have a single site, so these two problems, with four and eight sites and the disk Dirichlet-jump instances of the sampled suite, are the ones on which the simultaneous compilation of several terms can be checked against a known solution. \Cref{fig:before_after_arcs,fig:before_after_multi} show improvements of the same order in both cases, $1.7 \times 10^{5}$ and $1.2 \times 10^{5}$. Both runs carry a complex value-jump coefficient, so their error grows logarithmically as a jump is approached (\cref{app:holomorphic_evaluation}). We emphasize that this evidence is limited: both problems were run at a moderate budget and a single seed, both solutions lie in the span of the singular terms, neither is taken from the literature, and no case considered here exceeds eight sites, so the behaviour at larger site counts remains untested.

\FloatBarrier

\section{Ablation for the Holomorphic Network}
\label{app:holomorphic_ablation}

\Cref{tab:ablation} varies one factor at a time from the configuration the study started with, on the tuning suite of \cref{fig:suite_tuning}. The configuration used for \cref{fig:pairs} and the named benchmarks takes the best value found on the width, depth, learning-rate, Adam-step, and sample axes, so no single row of the table matches it; on the same suite it scores a ratio of $0.024$. Each of these values lies at the edge of the searched range, and doubling both the Adam steps and the points per segment lowers the ratio further, to $0.013$, so the budget used is not saturated. The number of terms and the exponent regime of each named benchmark (\cref{tab:benchmarks}) are set per problem, not by this ablation. Because the ratio is a median over six families, rows that act on only some families stay near one: training the exponents gives $0.34$ and $0.54$ on the two many-segment families, where no run converged, $0.97$ and $1.54$ on the mixed families, and exactly one on the Dirichlet-jump families, which carry no power terms; ladder depth 4 gives $0.49$ and $0.59$ on the square many-segment and mixed families but $17.5$ and $2.0$ on their disk counterparts.

\begin{table}[ht]
  \caption{Ablation of the enriched holomorphic model, one factor changed at a time from the starting configuration (width 32, depth 3, learning rate 0.003, 20\,000 Adam steps, 6\,000 L-BFGS steps, 512 points per segment, ladder depth 1). Every configuration was run on the tuning suite of \cref{fig:suite_tuning}, one instance per family with three seeds, which shares no instance with the evaluation suite. \emph{Boundary loss} is the median over the 18 runs. \emph{Ratio} is the median over the six families of this configuration's median boundary loss divided by the starting configuration's, so below one is an improvement. \emph{Converged} counts runs reaching a boundary loss below $10^{-6}$, and \emph{Time} is the median wall-clock per run. Trained exponents and ladder depth leave the two Dirichlet-jump families unchanged, which holds those rows near one; the text gives their per-family ratios. The plain row uses the starting configuration and was not tuned separately.}
  \label{tab:ablation}
  \centering
  \small
  \begin{tabular}{lrrrrr}
    \toprule
    Configuration & Runs & Boundary loss & Ratio & Converged & Time (s) \\
    \midrule
    starting configuration & 18 & $6.4 \times 10^{-5}$ & 1.00 & 3/18 & 402 \\
    \midrule
    depth 2 & 18 & $3.3 \times 10^{-6}$ & 0.03 & 7/18 & 280 \\
    learning rate 0.01 & 18 & $8.1 \times 10^{-6}$ & 0.14 & 7/18 & 381 \\
    Adam steps 40\,000 & 18 & $1.4 \times 10^{-5}$ & 0.41 & 5/18 & 644 \\
    width 16 & 18 & $1.4 \times 10^{-5}$ & 0.42 & 7/18 & 275 \\
    points per segment 1\,024 & 18 & $8.5 \times 10^{-6}$ & 0.43 & 6/18 & 645 \\
    depth 4 & 18 & $6.0 \times 10^{-5}$ & 0.84 & 1/18 & 512 \\
    L-BFGS steps 12\,000 & 18 & $3.8 \times 10^{-5}$ & 0.85 & 3/18 & 505 \\
    initialization empirical & 18 & $7.6 \times 10^{-5}$ & 0.95 & 3/18 & 381 \\
    model enriched, trained $\lambda$ & 18 & $5.1 \times 10^{-5}$ & 0.98 & 3/18 & 457 \\
    ladder depth 2 & 18 & $6.9 \times 10^{-5}$ & 1.00 & 3/18 & 415 \\
    ladder depth 4 & 18 & $8.0 \times 10^{-5}$ & 1.00 & 3/18 & 487 \\
    L-BFGS steps 3\,000 & 18 & $7.2 \times 10^{-5}$ & 1.13 & 3/18 & 325 \\
    learning rate 0.001 & 18 & $6.5 \times 10^{-5}$ & 1.38 & 4/18 & 385 \\
    Adam steps 10\,000 & 18 & $1.4 \times 10^{-4}$ & 2.33 & 1/18 & 254 \\
    width 64 & 18 & $1.4 \times 10^{-4}$ & 2.36 & 3/18 & 1\,079 \\
    Adam steps 1\,000 & 18 & $3.8 \times 10^{-4}$ & 3.54 & 4/18 & 138 \\
    Adam steps 3\,000 & 18 & $5.3 \times 10^{-4}$ & 7.07 & 2/18 & 165 \\
    points per segment 256 & 18 & $1.6 \times 10^{-3}$ & 11.39 & 3/18 & 238 \\
    model plain & 18 & $4.0 \times 10^{-3}$ & 60.97 & 0/18 & 369 \\
    \bottomrule
  \end{tabular}
\end{table}

\FloatBarrier

\section{Trained Against Frozen Exponents}
\label{app:lambda}

A trained exponent carries one real parameter, $\lambda=\epsilon+\operatorname{softplus}(\tau)$, initialised at its value in \cref{tab:dictionary}. The interior residual vanishes whatever the exponent, so $\lambda$ is driven by the boundary loss alone. Training makes the model tolerant of a wrong starting exponent but gives no consistent gain when the exponent is known, and on Motz's problem and the cracked beam the leading coefficient trades off against the exponent, which is why we pin the exponents there.

\Cref{tab:lambda} compares frozen and trained exponents. Within each pair, the two arms share the seed, and with it the initialization, training points, and evaluation points, except on Motz's problem and the cracked beam, whose frozen arm is the run of \cref{tab:all_models} and shares only the evaluation points. Starting at $10\%$ off, a frozen exponent is two to four orders of magnitude less accurate than one frozen at $\lambda^\ast$, and its boundary loss, $4.6\times10^{-7}$ to $3.0\times10^{-6}$ against at most $1.7\times10^{-11}$ in every other run of that block, showcases the error without a reference solution; the trained exponent recovers from either side to the accuracy of the correctly frozen one. Started at $\lambda^\ast$, freezing and training give the same accuracy within the spread over seeds on the four variants; only on the slit disk is every frozen run more accurate than every trained one. On Motz's problem and the cracked beam, training moves the leading exponent by $3\times10^{-6}$ to $8\times10^{-5}$, and the recovered $D_0$ has a relative error of $4.8$ to $7.4$ times that drift in all six runs, so the coefficient trades off against the exponent; the second exponent moves by $10^{-4}$ to $10^{-2}$ and the third and fourth by $0.10$ to $0.63$, so a trained ladder does not identify the higher terms of the expansion.

\begin{table}[ht]
  \caption{Frozen against trained exponents: median over seeds, with the range below it. The first two blocks start the exponent at its value $\lambda^\ast$ in \cref{tab:dictionary} and use the smallest budget of \cref{tab:ladder_seeds}, three seeds per arm; on Motz with a pole, all four ladder exponents are frozen or trained. The last block starts the exponent at $0.9\lambda^\ast$ and $1.1\lambda^\ast$ (six runs per arm) at the budget of \cref{tab:before_after}, against three runs frozen at $\lambda^\ast$. The last column gives the smallest and largest distance of the trained leading exponent from its exact value $\lambda_1^\ast$. ${}^{\ddagger}$The frozen arm is the run of \cref{tab:all_models}. ${}^{*}$$D_0$ read at the drifted $\lambda_1$.}
  \label{tab:lambda}
  \centering\scriptsize
  \setlength{\tabcolsep}{3pt}
  \begin{tabular}{@{}lcccc@{}}
    \toprule
    Problem & Frozen at $\lambda^\ast$ & Frozen, $10\%$ off & Trained & $|\lambda_1-\lambda_1^\ast|$ after training\\
    \midrule
    \multicolumn{5}{@{}l}{\emph{Started at $\lambda^\ast$; relative $L^2$ error}}\\
    Slit disk & $2.4{\times}10^{-11}$ & -- & $1.0{\times}10^{-10}$ & $8{\times}10^{-13}$--$6{\times}10^{-11}$\\
     & [$2.3{\times}10^{-11}$, $4.4{\times}10^{-11}$] & & [$5.9{\times}10^{-11}$, $1.8{\times}10^{-10}$] & \\
    L-shape $+\operatorname{Re}e^{z}$ & $1.2{\times}10^{-8}$ & -- & $1.5{\times}10^{-8}$ & $9.2{\times}10^{-8}$--$1.3{\times}10^{-6}$\\
     & [$8.6{\times}10^{-9}$, $3.1{\times}10^{-8}$] & & [$6.6{\times}10^{-9}$, $2.1{\times}10^{-8}$] & \\
    Pure mode $+\operatorname{Re}e^{z}$ & $1.7{\times}10^{-8}$ & -- & $1.4{\times}10^{-8}$ & $2.5{\times}10^{-8}$--$9.2{\times}10^{-7}$\\
     & [$1.3{\times}10^{-8}$, $2.5{\times}10^{-8}$] & & [$8.5{\times}10^{-9}$, $2.2{\times}10^{-8}$] & \\
    Sector $3\pi/2$ + pole & $8.6{\times}10^{-6}$ & -- & $9.3{\times}10^{-6}$ & $1.3{\times}10^{-5}$--$3.6{\times}10^{-4}$\\
     & [$7.0{\times}10^{-6}$, $1.3{\times}10^{-5}$] & & [$7.1{\times}10^{-6}$, $1.4{\times}10^{-5}$] & \\
    Motz + pole & $8.2{\times}10^{-6}$ & -- & $1.1{\times}10^{-5}$ & $5.7{\times}10^{-4}$--$1.1{\times}10^{-3}$\\
     & [$5.8{\times}10^{-6}$, $8.3{\times}10^{-6}$] & & [$7.1{\times}10^{-6}$, $1.9{\times}10^{-5}$] & \\
    \midrule
    \multicolumn{5}{@{}l}{\emph{Started at $\lambda^\ast$; relative error of $D_0$}}\\
    Motz$^{\ddagger}$ & $4.5{\times}10^{-6}$ & -- & $1.8{\times}10^{-4}$\rlap{${}^{*}$} & $2.5{\times}10^{-5}$--$8.2{\times}10^{-5}$\\
     & [$3.8{\times}10^{-6}$, $5.2{\times}10^{-6}$] & & [$1.6{\times}10^{-4}$, $4.8{\times}10^{-4}$] & \\
    Cracked beam$^{\ddagger}$ & $2.0{\times}10^{-6}$ & -- & $4.6{\times}10^{-5}$\rlap{${}^{*}$} & $3.0{\times}10^{-6}$--$7.9{\times}10^{-6}$\\
     & [$3.8{\times}10^{-7}$, $6.2{\times}10^{-6}$] & & [$1.5{\times}10^{-5}$, $4.6{\times}10^{-5}$] & \\
    \midrule
    \multicolumn{5}{@{}l}{\emph{Started $10\%$ off; relative $L^2$ error}}\\
    Pure mode & $3.4{\times}10^{-8}$ & $5.6{\times}10^{-4}$ & $3.1{\times}10^{-8}$ & $2.6{\times}10^{-9}$--$4.3{\times}10^{-7}$\\
     & [$2.2{\times}10^{-8}$, $4.8{\times}10^{-8}$] & [$4.9{\times}10^{-4}$, $6.5{\times}10^{-4}$] & [$7.8{\times}10^{-9}$, $1.0{\times}10^{-7}$] & \\
    Sector $3\pi/2$ & $4.6{\times}10^{-8}$ & $3.0{\times}10^{-4}$ & $2.7{\times}10^{-8}$ & $1.1{\times}10^{-8}$--$2.8{\times}10^{-7}$\\
     & [$3.2{\times}10^{-8}$, $7.6{\times}10^{-8}$] & [$2.4{\times}10^{-4}$, $4.0{\times}10^{-4}$] & [$9.9{\times}10^{-9}$, $3.4{\times}10^{-8}$] & \\
    Pure mode $+\operatorname{Re}e^{z}$ & $6.2{\times}10^{-8}$ & $2.2{\times}10^{-4}$ & $5.0{\times}10^{-8}$ & $3.6{\times}10^{-7}$--$1.7{\times}10^{-6}$\\
     & [$5.2{\times}10^{-8}$, $1.8{\times}10^{-7}$] & [$2.0{\times}10^{-4}$, $2.3{\times}10^{-4}$] & [$2.3{\times}10^{-8}$, $1.5{\times}10^{-7}$] & \\
    L-shape $+\operatorname{Re}e^{z}$ & $5.0{\times}10^{-7}$ & $1.8{\times}10^{-4}$ & $3.9{\times}10^{-7}$ & $4.0{\times}10^{-6}$--$1.4{\times}10^{-5}$\\
     & [$3.3{\times}10^{-7}$, $5.8{\times}10^{-7}$] & [$1.6{\times}10^{-4}$, $2.2{\times}10^{-4}$] & [$2.7{\times}10^{-7}$, $1.0{\times}10^{-6}$] & \\
    \bottomrule
  \end{tabular}
\end{table}

\end{document}